\documentclass[11pt,a4paper]{article}
\usepackage[margin=1in]{geometry}

\usepackage{amssymb}
\usepackage{selinput}
\usepackage{subfigure}
\usepackage{url}
\usepackage[normalem]{ulem}
\usepackage{amsmath}
\usepackage{ifthen}
\usepackage{color}
\usepackage{graphicx}
\usepackage{url}
\usepackage{soul}
\usepackage{color}
\usepackage{booktabs}       
\usepackage{multirow}
\usepackage{algorithm}
\usepackage{algpseudocode}
\usepackage{hyperref}
\usepackage{natbib}
\usepackage{authblk}

\DeclareRobustCommand{\old}[1]{}
\DeclareRobustCommand{\new}[1]{#1}

\begin{document}
\title{Structure-aware Bayesian full waveform inversion via diffusion-regularized Stein variational gradient descent}

\renewcommand{\thefootnote}{\fnsymbol{footnote}}

\author[1]{Chao Li}
\author[1]{Mrinal K. Sen}
\author[1]{Sergey Fomel}
\author[1]{Yangkang Chen}
\affil[1]{The University of Texas at Austin}

\maketitle
\begin{abstract}
Full waveform inversion (FWI) is a powerful tool for high-resolution subsurface model reconstruction, but it remains highly nonlinear and ill-posed, especially when the initial model is inaccurate or the acquired data are noisy and incomplete. Annealed Stein variational gradient descent (ASVGD) provides a particle-based Bayesian framework for FWI by approximating the posterior distribution with an ensemble of velocity models, thereby enabling uncertainty quantification beyond a single deterministic solution. However, conventional ASVGD-FWI commonly relies on simple handcrafted priors, such as Gaussian or smoothing assumptions, which are often insufficient to represent complex geological structures. To address this limitation, we introduce a learned diffusion model (DM) as an implicit regularization mechanism within the ASVGD-FWI framework. Instead of explicitly evaluating the diffusion prior density, the trained diffusion model acts as a learned geological projection operator that guides updated particles toward the distribution of plausible velocity models. In this way, the proposed method combines data consistency from FWI, posterior diversity from ASVGD, and structural realism from the diffusion prior. Numerical experiments demonstrate that the diffusion-regularized ASVGD-FWI produces more geologically consistent mean models, suppresses nonphysical artifacts, and yields uncertainty estimates that are more concentrated around structurally complex and poorly constrained regions. These results suggest that learned diffusion priors can effectively enhance Bayesian FWI by improving both inversion stability and the interpretability of posterior uncertainty.
\end{abstract}


\section{Introduction}
\par Full waveform inversion (FWI) has been attracting more and more attention in the field of seismic exploration due to its superior capability of high-resolution subsurface model construction \citep{pratt1998,virieus2009}. It can refine the subsurface models by iterative minimizing the differences between the observed data and the predicted data \citep{vantassel2022,lic2022jge,dhara2023elastic}. However, FWI is a highly nonlinear inverse problem, and an inaccurate initial model may cause the inversion to converge to an erroneous solution \citep{gauthier1986, agfwi2012, tariq2016fwi}. Such a convergence failure can further degrade subsequent seismic imaging and interpretation. 

\par To address this issue, considerable efforts have been made to stabilize the inversion process from different perspectives. One common strategy is to construct a reliable initial model using traveltime tomography or prior information from well logs \citep{tailland2009,yaojs2022,zhangqc2025}, which can effectively reduce background velocity errors and provide a more favorable starting point for inversion. Next, a variety of robust objective functions have been developed to enlarge the convergence basin of FWI and alleviate the limitations of the commonly used L2-norm misfit \citep{warner2016,zhu2016building,yangyn2018,sunbb2022,Licqawi2024}. In addition, low-frequency information is known to reduce FWI's dependence on the accuracy of the initial model, and reconstructing missing low-frequency components has therefore become an effective alternative strategy \citep{oleg2019,Wangzy2022,chengsj2024low}. Recently, motivated by the remarkable success of deep learning (DL) in modeling complex nonlinear mappings, data-driven FWI (DDFWI) has been proposed to directly estimate subsurface models from shot gathers without requiring an initial model \citep{yangfs2019,shucai2020,zhangzp2020,dhara2023elastic,mahdi2024}. This strategy has the potential to substantially advance FWI by alleviating its strong dependence on accurate initial models.

\par However, most of the aforementioned approaches rely on deterministic formulations. As a result, they typically yield only a single possible solution and cannot provide a quantitative assessment of the reliability or uncertainty of the inverted model. To address this limitation, increasing attention has been paid to Bayesian inference (nondeterministic FWI frameworks), which aim to estimate not only the posterior mean but also the associated posterior variance \citep{Sambridge2002,sen2013global}. The central idea of the Bayesian method is to estimate the posterior probability distribution over all possible models that can explain the observed data, thereby providing a more comprehensive characterization of the solution space. Generally, Markov chain Monte Carlo (MCMC) is one of the most classical approaches for characterizing the posterior distribution, in which a sequence of samples is generated by exploring the model space through stochastic transitions \citep{MartinJ2012,def2019}. Within the MCMC framework, many successful applications have been demonstrated in geophysical inverse problems, such as gravity inversion \citep{Mosegaard1995}, resistivity inversion \citep{Malinverno2002}, dispersion inversion \citep{bodint2012,galettie2017}, and FWI \citep{raya2018,fuxin2021,hush2025}. Nevertheless, conventional MCMC methods are often computationally demanding, which limits their applicability to high-dimensional problems \citep{corm2025}. Consequently, more efficient MCMC variants have been introduced, including delayed-acceptance algorithms \citep{Christen01122005,cuit2011}, gradient-based MCMC \citep{Aleardi2021,zhaozy2022,Berti2024}, and Hamiltonian Monte Carlo \citep{girolami2011,senmrinal2017,Fichter2019}.

\par More recently, the rapid progress of generative models, especially diffusion models, has inspired the use of learned diffusion priors to enhance the structural fidelity of inversion results, leading to encouraging performance \citep{wangf2023,wang2025wavediffusion,lic2026}. Inspired by the capability of diffusion models to characterize complex model distributions, \cite{taufik2026} introduced a diffusion-model-based particle initialization scheme into the SVGD framework, thereby improving the computational efficiency and robustness of conventional SVGD-FWI. To further investigate the role of structural prior information in conventional SVGD-FWI, we propose a \old{diffusion-prior-regularized}\new{diffusion-regularized} SVGD framework for uncertainty evaluation in FWI. The learned diffusion prior is trained on a newly developed dataset, GeoFWI \citep{lcgeowi2026}, which contains more geologically realistic structural patterns. Instead of using the diffusion model only for particle initialization or post-processing, we incorporate the pretrained diffusion model as an implicit regularization term directly into the SVGD particle update. In this way, the particle evolution is jointly guided by the data likelihood and the learned geological prior. 

\par Unlike the classic SVGD framework, where particle updates are mainly driven by the data likelihood, likelihood-driven updates alone may produce geologically inconsistent samples and inversion artifacts, especially in the presence of noise, limited illumination, or data incompleteness. By incorporating a learned diffusion prior, the particles are guided toward realistic subsurface model distributions during the update process. This helps enhance structural fidelity, stabilize the inversion, and improve the reliability and interpretability of posterior uncertainty estimation. Numerical experiments on both synthetic and field data demonstrate that the proposed method significantly reduces posterior uncertainty and produces more geologically reasonable inversion results, confirming the effectiveness and superiority of the diffusion-regularized SVGD-FWI framework.

\section{THEORY}
\subsection{GeoFWI dataset and diffusion prior}
\begin{figure}[htb!]
\centering
\includegraphics[width=1\columnwidth]{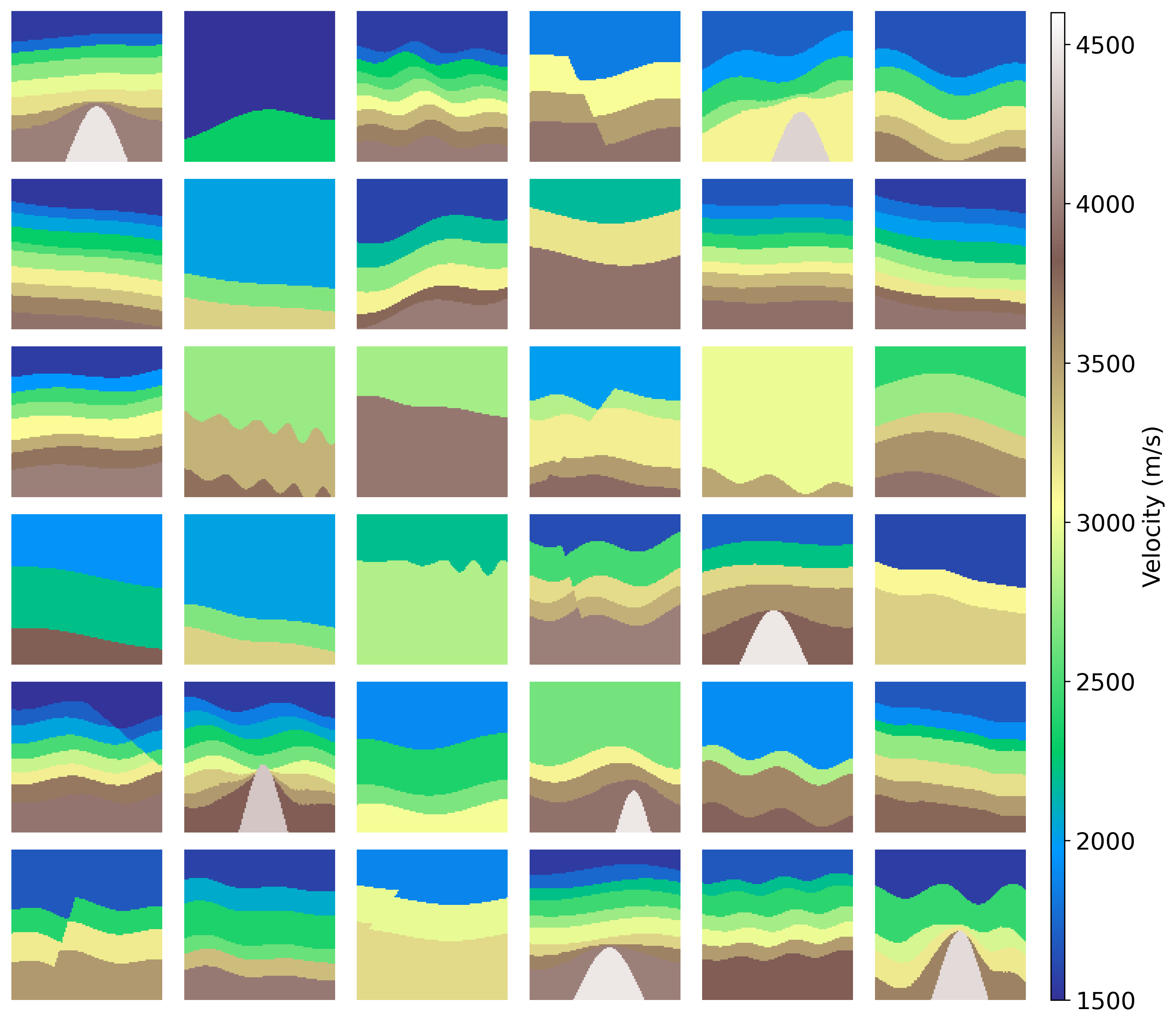}
\caption{Representative samples from the GeoFWI dataset, including folding structures, faults, and salt bodies with varying numbers of layers.}
\label{fig:geofwigal}
\end{figure}

We begin this section by briefly describing the GeoFWI dataset, which serves as the training dataset for learning the diffusion prior. The recently developed GeoFWI dataset contains approximately 50,000 velocity models that cover a variety of geological structures, including folds, faults, and salt bodies. Each velocity model is represented on a (100 $\times$ 100) grid, and the velocity values span from 1500 m/s to 4500 m/s. The dataset is constructed with progressively increasing structural complexity, where the number of layers ranges from 2 to 11. The dataset can be accessed at \url{https://github.com/aaspip/geofwi}, and the detailed sample counts for each model family are provided in \citep{lcgeowi2026}. As shown in Figure~\ref{fig:geofwigal}, the GeoFWI dataset contains representative velocity models with different geological patterns and structural complexities. These examples involve various subsurface scenarios, including folded layers, faulted structures, salt bodies, and models with different numbers of stratigraphic layers. The dataset, therefore, captures both large-scale background velocity variations and local high-contrast anomalies, providing rich structural information for learning the distributions of geological models. This structural diversity is particularly important for diffusion prior to training, as it enables the pretrained diffusion model to characterize realistic subsurface patterns and guide the SVGD particles toward geologically plausible solutions during inversion.

\par Next, a denoising diffusion probabilistic model (DDPM) \citep{ho2020denoising} is employed to learn an implicit structural prior from the training velocity models. The forward diffusion process gradually corrupts clean samples by adding Gaussian noise over multiple time steps (e.g., $[t_{0},t_{\mathrm{T}}]$) until they approach a nearly pure noise distribution. Conversely, the reverse process trains a neural denoising network to progressively remove the injected noise and reconstruct samples that follow the learned data distribution. Mathematically, given a clean sample $\mathbf{m}_{0}\sim p(\mathbf{m})$, the forward and reverse diffusion process can be expressed using a stochastic differential equation (SDE) \citep{song2020score} 
\begin{equation}
dm_t = -\frac{\eta(t)}{2}m_tdt + \sqrt{\eta(t)}d\omega_t,
\label{eq:diffusionfor}
\end{equation}
\begin{equation}
dm_t = \left[-\frac{\eta(t)}{2}m_t - \eta(t) \nabla_{m_t} \log p_t(m_t)\right]dt + \sqrt{\eta(t)}d\hat{\omega}_t.
\label{eq:diffusionrev}
\end{equation}
where $\eta(t)$ is the noise schedule, $m_t$ is the model status at time $t$, $\omega_t\sim q(\mathbf{m})$ is a predefined Brownian motion, and the score function $\nabla_{m_t} \log p_t(m_t) $ represents the data log probability density at time $t$. \new{Specifically, we employ a U-Net architecture with a base channel width of 64 and a single input/output channel. The diffusion process consists of 1000 forward steps, while 250 denoising diffusion implicit model (DDIM) sampling steps are used during inference. The model is trained using the standard noise-prediction objective with a batch size of 32 and a learning rate of \(8\times10^{-5}\) for 300000 iterations. We adopt an exponential moving average of the model parameters with a decay rate of 0.995 to improve training stability; more details are provided in \citep{lcgeowi2026,yiran2026}.}

\subsection{Theory of annealed SVGD FWI}
\par Bayesian inference aims to estimate the posterior probability density function (PDF) $p(\mathbf{m}\mid \mathbf{d}_{\mathrm{obs}})$ conditioned on the observed data $\mathbf{d}_{\mathrm{obs}}$. According to the Bayes's rule, \new{the PDF $p(\mathbf{m}\mid \mathbf{d}_{\mathrm{obs}})$ can be written as
\begin{equation}
p(\mathbf{m}\mid \mathbf{d}_{\mathrm{obs}})=\frac{p(\mathbf{d}_{\mathrm{obs}}\mid \mathbf{m})p(\mathbf{m})}{p(\mathbf{d}_{\mathrm{obs}})},
\label{eq:bays}
\end{equation}
\begin{equation}
p(\mathbf{d}_{\mathrm{obs}}\mid \mathbf{m})\propto \exp\left[\frac{1}{2\sigma_d^2}\|\mathbf{G}(\mathbf{m})-\mathbf{d}_{\mathrm{obs}}\|_2^2\right],
\label{eq:datap}
\end{equation}
where $p(\mathbf{d}_{\mathrm{obs}}\mid \mathbf{m})$ represents the likelihood, which quantifies how well the synthetic data generated from model $\mathbf{m}$ explains the observed data $\mathbf{d}_{\mathrm{obs}}$, and $\sigma_d^2$ is the Gaussian noise variance.}  Moreover, $p(\mathbf{m})$ represents the prior information imposed on the model parameters, and $p(\mathbf{d}_{\mathrm{obs}})$ is the normalization constant, which is also called the evidence.

\par Due to the substantial computational cost and complexity associated with direct sampling from the posterior distribution $p(\mathbf{m}\mid\mathbf{d}_{\mathrm{obs}})$, SVGD has emerged as an efficient deterministic particle-based inference method. Starting from an initial surrogate distribution $q(\mathbf{m})$, SVGD iteratively transports particles toward the target posterior distribution $p(\mathbf{m}\mid \mathbf{d}_{\mathrm{obs}})$ by seeking an optimal transformation that minimizes the KL  divergence between $q(\mathbf{m})$ and $p(\mathbf{m}\mid \mathbf{d}_{\mathrm{obs}})$, \new{which has the following form
\begin{equation}
\mathrm{KL}(q(\mathbf{m})\mid \mid p(\mathbf{m}\mid \mathbf{d}_{\mathrm{obs}}))=\mathrm{E}_{\mathbf{m}\sim q}[-\log p(\mathbf{m}\mid \mathbf{d}_{\mathrm{obs}})+\log q(\mathbf{m})].
\label{eq:kld}
\end{equation}}
Within the SVGD framework \citep{liuq2019}, the optimal update direction ($\phi^*(\mathbf{m})$) is derived from the gradient of the KL divergence, guiding the approximate distribution $q(\mathbf{m})$ toward the target posterior distribution $p(\mathbf{m}\mid \mathbf{d}_{\mathrm{obs}})$. Given a set of initial particles, \new{the gradient $\phi^*$ can be derived as
\begin{equation}
\phi^*(\mathbf{m})=\mathrm{E}_{\mathbf{m}\sim q}
[\underbrace{k(\mathbf{m},\cdot)\nabla_{\mathbf{m}}\log p(\mathbf{m}\mid\mathbf{d}_\mathrm{obs})}_{\mathrm{driving\ force}}+\underbrace{\nabla_{\mathbf{m}}k(\mathbf{m},\cdot)}_{\mathrm{repulsive\ force}}],
\label{eq:klgrad}
\end{equation}}
\begin{equation}
k(\mathbf{m},\hat{\mathbf{m}})=\exp \left[-\frac{\|\mathbf{m}-\hat{\mathbf{m}}\|^2}{2h^2}\right],
\label{eq:kernel}
\end{equation}
where $k(\cdot,\cdot)$ denotes the kernel function. In this study, we adopt a radial basis function (RBF) kernel to quantify particle similarity and govern their interactions during updates. \new{The parameter $h$ denotes the kernel bandwidth, which controls the effective range of particle interactions and the kernel's sensitivity to interparticle distances. Commonly, we can set $h=\hat{d}^2/log N$, where $N$ denotes the total number of particles. Further details regarding the bandwidth selection strategy can be found in \citep{liuq2019}}. Reading from left to right, Equation~\ref{eq:klgrad} shows that the particle update direction is jointly determined by a data-misfit-driven driving force and a kernel-induced repulsive force associated with particle similarity. The former guides the particles toward regions of high posterior probability, while the latter preserves particle diversity, enabling the inversion to characterize posterior uncertainty rather than yielding only a single deterministic solution. 

\par Mathematically, since we are using a set of initial particles $\mathbf{m}^0\sim q_0(\mathbf{m})$ to approximate the expectation of equation~\ref{eq:klgrad}, it can be rewritten using the empirical average over the particles as 
\begin{equation}
\phi^*_{l}(\mathbf{m})=\frac{1}{N}\sum_{j=1}^{N}[k(\mathbf{m}_j^l,\mathbf{m})\nabla_{\mathbf{m}_{j}^{l}} \log p(\mathbf{m}_j^l\mid \mathbf{d}_{\mathrm{obs}})+\nabla_{\mathbf{m}_{j}^{l}} k(\mathbf{m}_j^l,\mathbf{m})],
\label{eq:klgrad1}
\end{equation}
where $\mathbf{m}_j$ represents the j-th particle, and $l$ is the current iteration index\old{and $\alpha$ is the step size}. However, when applied to high-dimensional inverse problems, classic SVGD can exhibit mode collapse and insufficient posterior dispersion, which may compromise uncertainty characterization. To mitigate these issues, \cite{corm2025} proposed an annealed SVGD (ASVGD) FWI framework in which a heuristic temperature parameter $\beta(l)\in[0,1]$ is introduced to adaptively balance the driving and repulsive forces throughout the iterations. The resulting annealed update direction $\hat{\phi}_{l}^{*}(\mathbf{m})$ is given by
\begin{equation}
\hat{\phi}^*_{l}(\mathbf{m})=\frac{1}{N}\sum_{j=1}^{N}[\beta(l)k(\mathbf{m}_j^l,\mathbf{m})\nabla_{\mathbf{m}_{j}^{l}} \log p(\mathbf{m}_j^l\mid \mathbf{d}_{\mathrm{obs}})+\nabla_{\mathbf{m}_{j}^{l}} k(\mathbf{m}_j^l,\mathbf{m})],
\label{eq:klgrad1a}
\end{equation}
\begin{equation}
\beta(l)=\tanh \left[\left(1.3\frac{l}{L}\right)^p\right],
\label{eq:betal}
\end{equation}
where $\beta(l)$ is hyperbolic tangent schedule \citep{dangelo2021}, \new{$L$ denotes the maximum number of iterations, and $p$ is empirically set to 3 \citep{corm2025}}. Then the initial particles can be updated via 
\begin{equation}
\mathbf{m}_j^{l+1}=\mathbf{m}_j^{l}+\alpha \hat{\phi}^*_{l}(\mathbf{m}_j^l).
\label{eq:klup}
\end{equation}
Compared with classic SVGD, annealed SVGD enables particles to explore a broader region of the model space during the early iterations and gradually concentrate in high-posterior-probability regions at later stages. This progressive transition helps mitigate premature particle collapse and better preserve posterior diversity.

\begin{figure}[htb!]
\centering
\includegraphics[width=1\columnwidth]{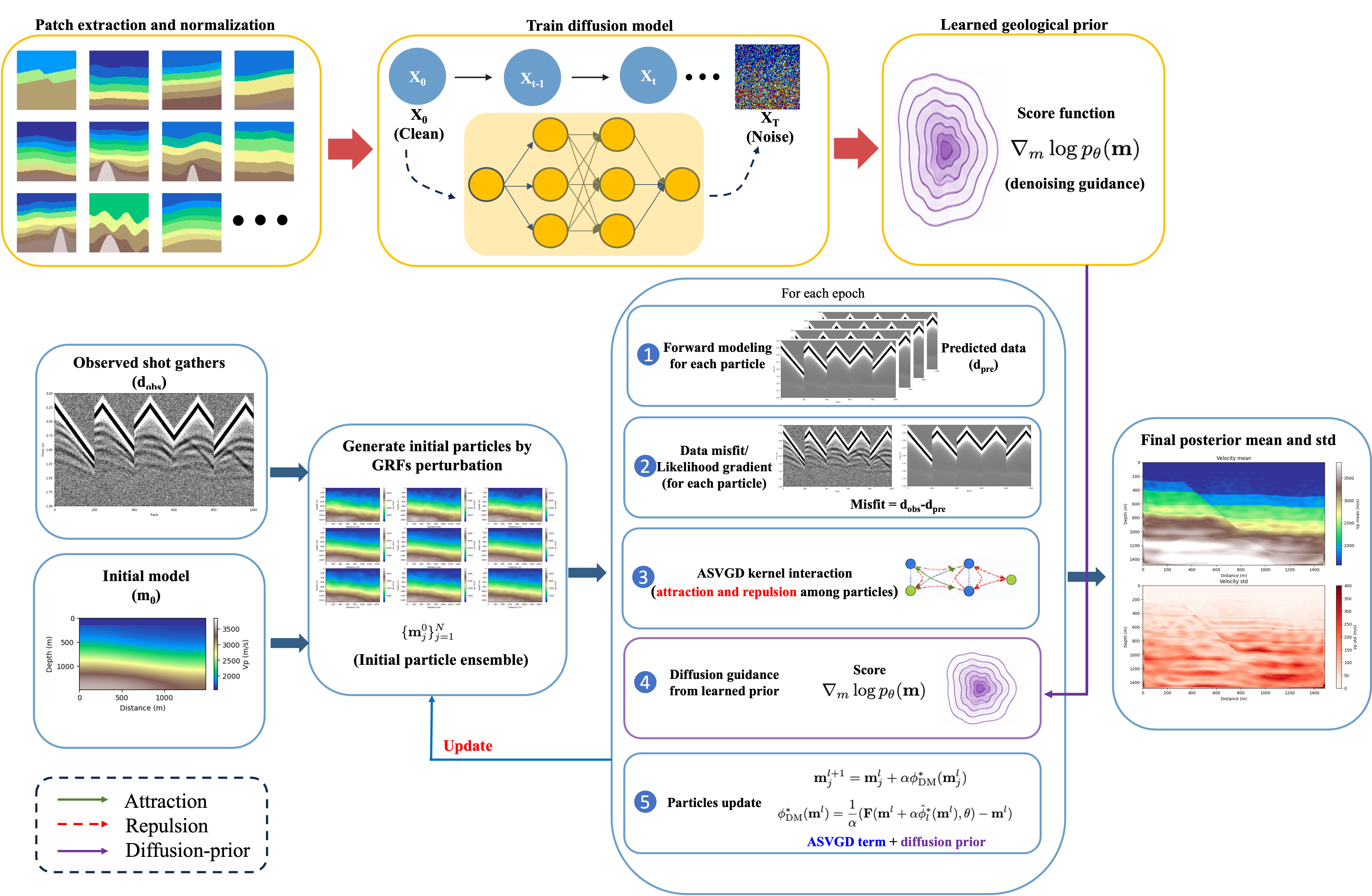}
\caption{A sketch of diffusion model guided annealed SVGD FWI, where a diffusion prior is introduced for improved geological plausibility.}
\label{fig:svgdfwi}
\end{figure}

\subsection{Annealed SVGD for FWI with implicit diffusion prior}
\par Although annealed SVGD provides an efficient framework for posterior approximation and uncertainty quantification, the conventional inversion process does not explicitly enforce the structural plausibility of the recovered models. In the presence of noise, limited acquisition coverage, or inaccuracies in the physical modeling, likelihood-driven particle updates may introduce nonphysical structures and inversion artifacts into the posterior samples, leading to biased estimates and unreliable uncertainty characterization. Therefore, incorporating a learned structural diffusion prior is essential to constrain the particle evolution toward geologically plausible regions of the model space, suppress spurious features, and improve both the robustness of the inversion and the credibility of the resulting uncertainty estimates. \new{Following the Bayesian formulation, diffusion regularization can be incorporated into the model prior distribution. Let $\mathbf{R}_{\mathrm{DM}}(\cdot)$ denote an implicit diffusion-model-based regularization functional that incorporates learned structural prior information into the inversion. Then the diffusion regularized model prior can be expressed as }
\begin{equation}
p_{\mathrm{DM}}(\mathbf{m})=\frac{1}{Z_{prior}}p(\mathbf{m})\exp[-\mathbf{R}_{\mathrm{DM}}(\mathbf{m})],
\label{eq:diffmodel}
\end{equation}
where $Z_{prior}$ is a normalization factor. According to Bayes' theorem, the corresponding posterior distribution with diffusion regularization is

\begin{equation}
\begin{aligned}
\hat{p}(\mathbf{m}\mid\mathbf{d}_{\mathrm{obs}})
&= \frac{p(\mathbf{d}_{\mathrm{obs}}\mid\mathbf{m})
p_{\mathrm{DM}}(\mathbf{m})}
{p(\mathbf{d}_{\mathrm{obs}})} \\
&= \frac{1}{Z_{\mathrm{prior}}}
p(\mathbf{m}\mid\mathbf{d}_{\mathrm{obs}})
\exp[-\mathbf{R}_{\mathrm{DM}}(\mathbf{m})].
\end{aligned}
\label{eq:diffusiondis}
\end{equation}

\begin{figure}[htb!]
\centering
\includegraphics[width=1.0\columnwidth]{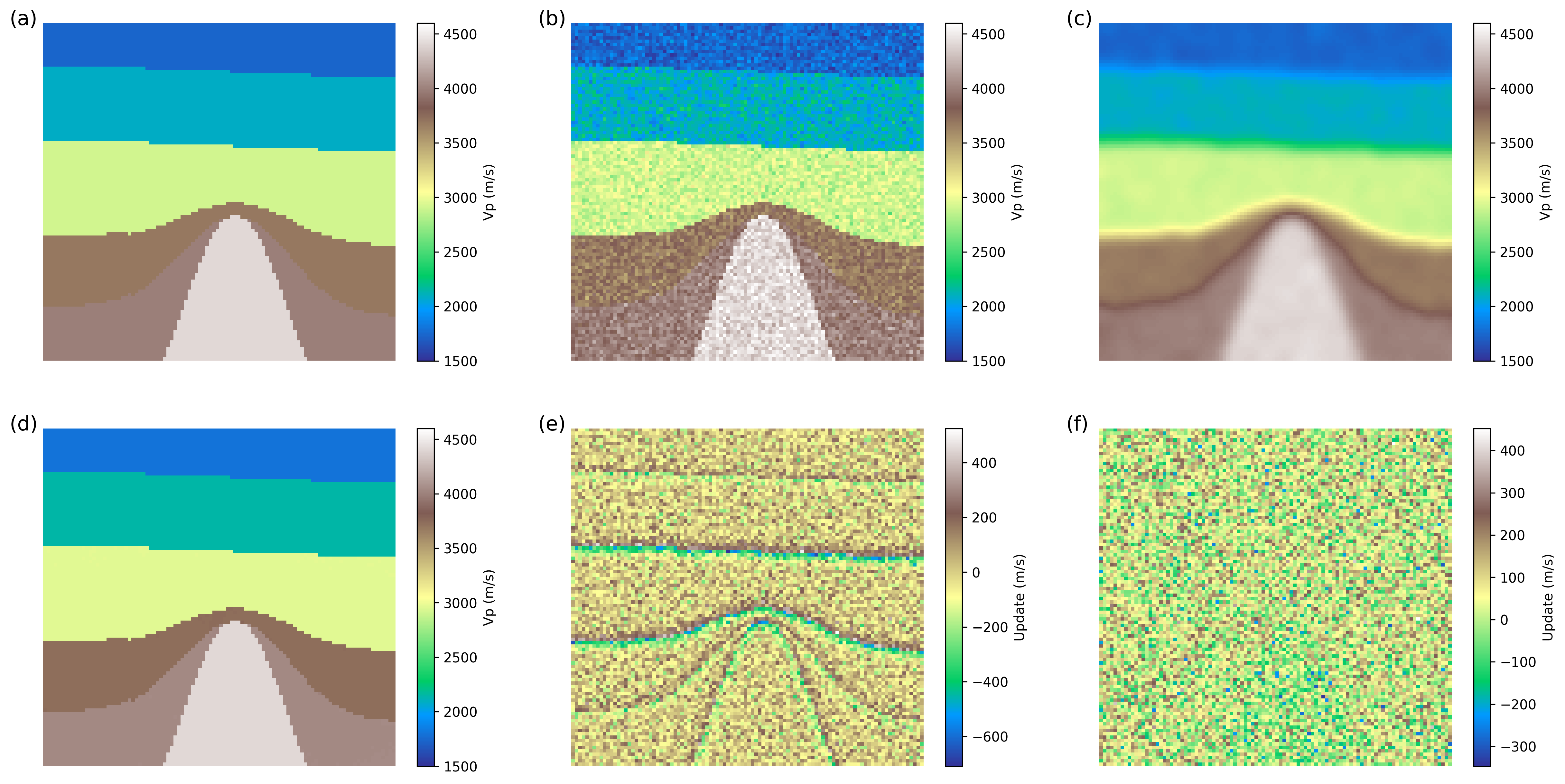}
\caption{Illustration of the structure-preserving behavior of the learned diffusion prior. (a) Clean velocity model. (b) Perturbed velocity model with random perturbation. (c) Filtered result using Gaussian smoothing. (d) Filtered result using the learned diffusion prior. (e) Filtering difference corresponding to Gaussian smoothing. (f) Filtering difference corresponding to the diffusion prior. Compared with Gaussian smoothing, the diffusion prior suppresses random perturbations while better preserving coherent geological structures and sharp interfaces. }
\label{fig:dmprior}
\end{figure}

\new{Taking the negative logarithm of equation~\ref{eq:diffusiondis}, we obtain
\begin{equation}
\log\hat{p}(\mathbf{m}\mid\mathbf{d}_{\mathrm{obs}})=\log p(\mathbf{m}\mid\mathbf{d}_{\mathrm{obs}})-\mathbf{R}_{\mathrm{DM}}(\mathbf{m})-\log Z_{post},
\label{eq:logdiffusiondis}
\end{equation}
where $Z_{post}$ is a normalization factor for posterior distribution. 
}
According to equation~\ref{eq:kld}, we reformulate the objective by incorporating a structural prior into the KL-divergence minimization, thereby constraining the posterior approximation toward geologically plausible solutions and improving the reliability of the final inversion results. \old{The diffusion prior guided KL divergence can be formulated as}\new{The diffusion prior guided variational objective function can be formulated as
\begin{equation}
\mathbf{J}(q)=\mathrm{KL}(q(\mathbf{m})\mid\mid \hat{p}(\mathbf{m}\mid\mathbf{d}_{\mathrm{obs}})).
\label{eq:kldiffusion}
\end{equation}
} 
\new{Using the definition of the KL divergence and substituting equation~\ref{eq:logdiffusiondis} into equation~\ref{eq:kldiffusion}, the variational objective function becomes
\begin{equation}
\mathbf{J}(q)=\mathrm{KL}(q(\mathbf{m})\mid\mid \hat{p}(\mathbf{m}\mid\mathbf{d}_{\mathrm{obs}}))-\log Z.
\label{eq:newobj}
\end{equation}
Since $Z$ is independent of $q(\mathbf{m})$, minimizing equation~\ref{eq:newobj} is equivalent to minimizing the KL divergence between $q(\mathbf{m})$ and the \new{diffusion-regularized} target distribution $\hat{p}(\mathbf{m}\mid\mathbf{d}_{\mathrm{obs}})$.}
To illustrate the effect of the learned diffusion prior, we conduct a simple smoothing test on a randomly selected velocity model. Figure~\ref{fig:dmprior}a shows the clean velocity model, and Figure~\ref{fig:dmprior}b shows the perturbed model after adding random noise. For comparison, we apply both Gaussian filtering (e.g., with a standard deviation of 2 for the Gaussian kernel) and the learned diffusion prior to the perturbed model. As shown in Figure~\ref{fig:dmprior}c, Gaussian filtering can suppress part of the random noise, but it also blurs sharp interfaces and smooths important geological structures. In contrast, the diffusion-regularized result in Figure~\ref{fig:dmprior}d better preserves the main layered structures and the salt-body boundary while reducing the random perturbations. Figures~\ref{fig:dmprior}e and \ref{fig:dmprior}f show the corresponding update fields generated by Gaussian filtering and the diffusion prior, respectively. The Gaussian-filter update exhibits strong responses along geological interfaces, indicating that the smoothing operator blurs not only random noise but also structural boundaries. By contrast, the diffusion-prior update is more spatially adaptive and less destructive to coherent geological structures. \new{This comparison suggests that the learned diffusion prior acts as an effective structure-preserving regularization mechanism, resulting in a lower mean squared error (MSE) of 12554.57 compared with 17590.99 obtained using Gaussian filtering.}. Then, following equation~\ref{eq:klgrad}, we can employ the shaping regularization framework \citep{fomel2007} to integrate the annealed SVGD update with the learned structural prior. \new{Accordingly, the diffusion-model-guided update direction, denoted by $\phi_{\mathrm{DM}}^*$, can be expressed as
\begin{equation}
\phi^*_{\mathrm{DM}}(\mathbf{m}^l)=\frac{1}{\alpha}(\mathbf{F}(\mathbf{m}^l+\alpha \hat{\phi}^*_l(\mathbf{m}^l),\theta)-\mathbf{m}^l),
\label{eq:kldiffusiongrad1}
\end{equation}
\begin{equation}
\mathbf{F}(\mathbf{x};\theta)\approx\operatorname{prox}_{\alpha R_{\mathrm{DM}}}(\mathbf{x})
=\arg\min_{\mathbf{z}}
\left[
\frac{1}{2\alpha}
\left\|\mathbf{z}-\mathbf{x}\right\|_2^2+R_{\mathrm{DM}}(\mathbf{z})
\right].
\end{equation}
where $\mathbf{F}(\mathbf{x};\theta)$ denotes the diffusion-based shaping operator that embeds the learned structural prior into the input $\mathbf{x}$, with $\theta$ representing the pretrained diffusion-model parameters. 
The particles can be refined iteratively with a diffusion prior via 
\begin{equation}
\mathbf{m}_j^{l+1}=\mathbf{m}_j^{l}+\alpha\phi^*_{\mathrm{DM}}(\mathbf{m}_j^l).
\label{eq:kldiffusionupdate}
\end{equation}}
\new{Based on equation~\ref{eq:kldiffusiongrad1}, we can recognize that the diffusion-guided ASVGD update follows a splitting scheme where the particles are first refined using the annealed SVGD gradient and then fully modified by the pretrained diffusion model, which serves as an implicit structural-shaping operator, projecting the particles toward the learned geological model manifold}. Specifically, Figure~\ref{fig:svgdfwi} illustrates the workflow of the proposed diffusion-regularized annealed SVGD inversion framework. After the annealed SVGD update, the resulting particle models are passed through the pretrained diffusion model, which serves as an implicit structural-shaping operator to enhance their geological plausibility.

\section{EXAMPLES}
In this section, we use several synthetic and field examples to validate the effectiveness of the proposed diffusion-regularized annealed SVGD FWI, which demonstrates superior stability and accuracy compared to the classic annealed SVGD method. Notably, the GeoFWI training samples have a fixed size of (100 $\times$ 100), which may differ from the dimensions of the velocity models used in subsequent numerical experiments. To address this size inconsistency when applying the diffusion prior, we adopt a commonly used patch-based strategy \citep{wangf2023,lic2026}. Specifically, in the following numerical examples, each velocity model is divided into overlapping (100 $\times$ 100) patches with a shift size of 10 to mitigate patch boundary artifacts. The diffusion prior is then applied to each patch independently, and the regularized patches are finally merged to reconstruct the complete \new{diffusion-regularized} model. \new{For seismic forward modeling, we use the Deepwave package \citep{richardson_alan_2026} to numerically solve the acoustic wave equation, and the inversion is implemented under the PyTorch framework with automatic differentiation.}

\subsection{Synthetic data}
\begin{figure}[htb!]
\centering
\includegraphics[width=1.0\columnwidth]{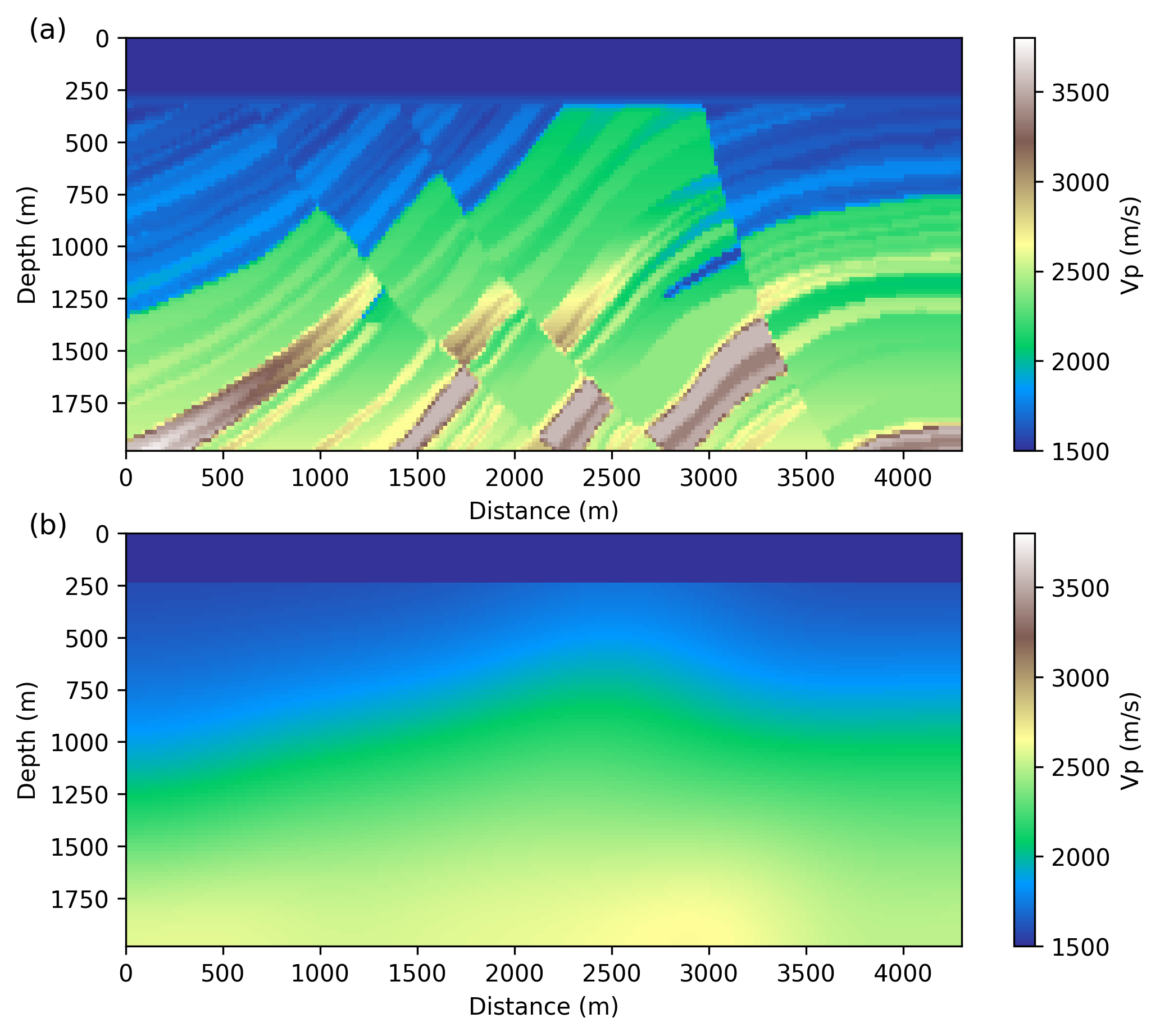}
\caption{(a) Marmousi model (ground truth). (b) The initial Marmousi model was used for inversion. }
\label{fig:marmini}
\end{figure}

\begin{figure}[htb!]
\centering
\includegraphics[width=1.0\columnwidth]{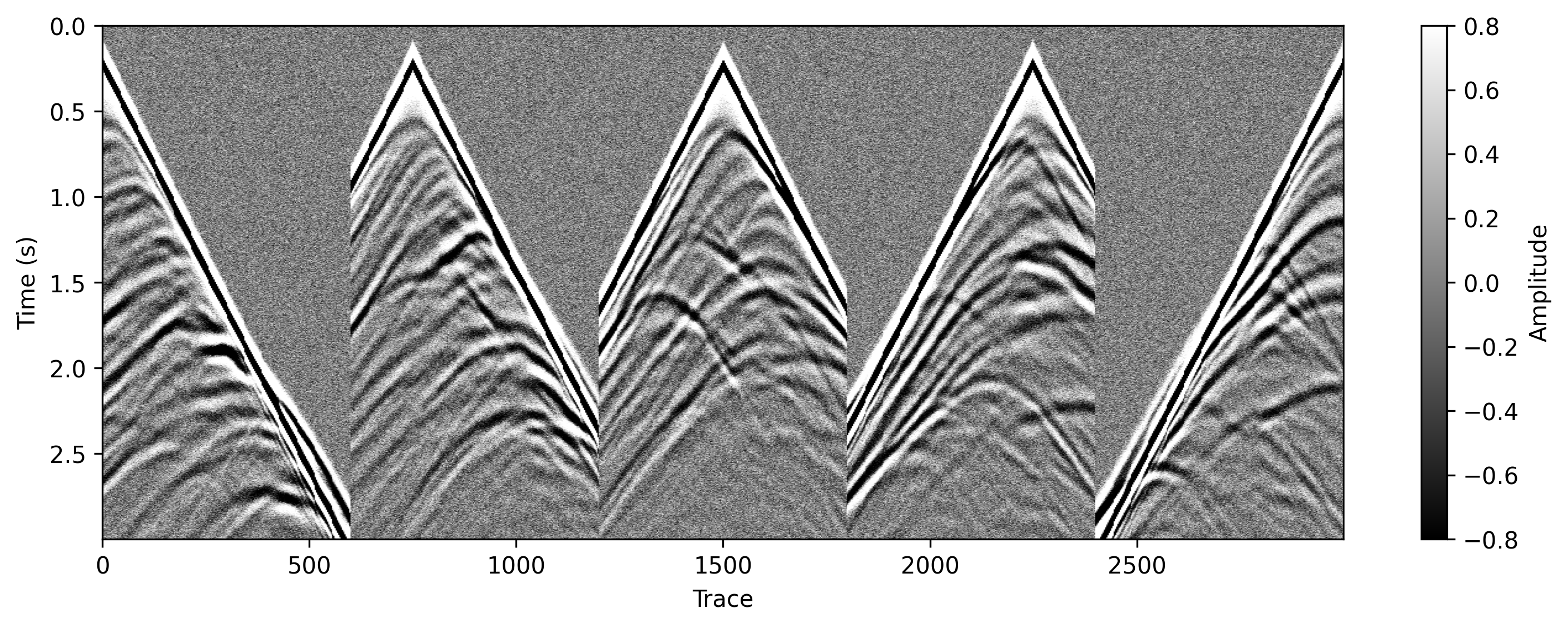}
\caption{Observed data generated using real Marmousi model (DSNR=15.7 dB). }
\label{fig:marmdata}
\end{figure}

\begin{figure}[htb!]
\centering
\includegraphics[width=1.0\columnwidth]{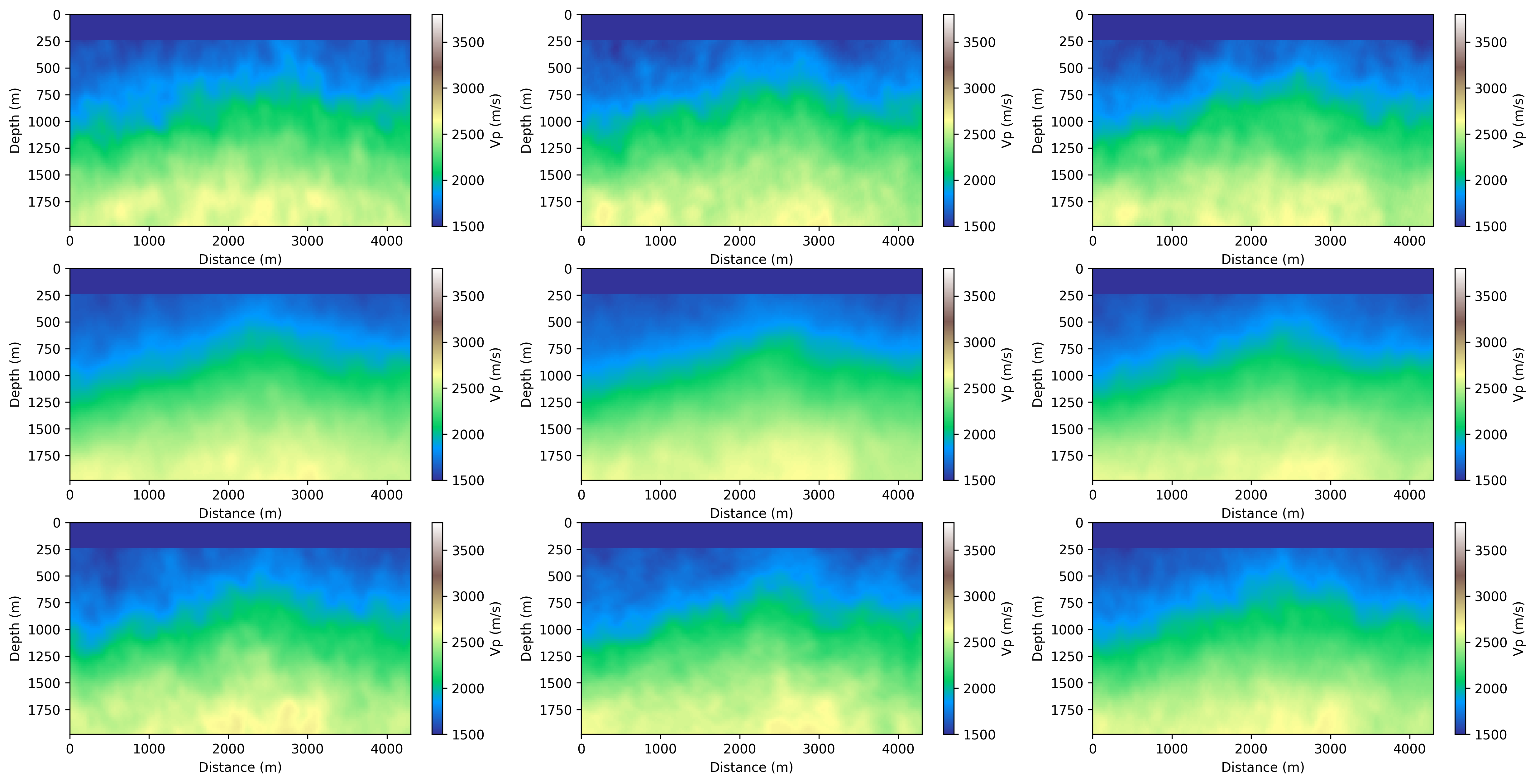}
\caption{Nine randomly selected initial particles generated by perturbing the smoothed Marmousi model with GRFs.}
\label{fig:marmpar}
\end{figure}

The first experiment is the well-known Marmousi model borrowed from \citep{corm2025} as shown in Figure~\ref{fig:marmini}. \new{The model consists of $100 \times 216$ grid points with a spatial interval of 20 m}. Figure~\ref{fig:marmini}a shows the ground-truth velocity model, while its smoothed counterpart, displayed in Figure~\ref{fig:marmini}b, is used as the initial model for inversion. A 7 Hz Ricker wavelet with a temporal sampling interval of 1 ms is employed as the source signature, and 5 shots are evenly distributed along the surface. \new{In addition, a water layer is imposed at the top of the model and remains fixed, and we set four absorptive boundary conditions throughout the inversion}. \old{Figure~\ref{fig:marmdata} shows the corresponding observed data with ambient noise}\new{Notably, we assume that the observed seismic data are contaminated by additive, independent Gaussian noise. Figure~\ref{fig:marmdata} shows the corresponding noisy observed data}. To evaluate the noise level of the observed data, \new{we define the data signal-to-noise ratio (DSNR) as
\begin{equation}
\mathrm{DSNR}=20\log_{10}\left(\left \| \mathbf{S}_{\mathrm{clean}}\right \|_{2}/\left \| \mathbf{N}_{\mathrm{noi}}\right \|_{2}\right),
\label{eq:sdnr}
\end{equation}
where $\mathbf{S}_{\mathrm{clean}}$ represents the clean seismic data, $\mathbf{N}_{\mathrm{noi}}$ is noise and the DSNR of Figure~\ref{fig:marmdata} is 15.7 dB}. \new{To construct the initial particle ensemble for the Marmousi inversion, Gaussian random fields (GRFs) are added to the smoothed initial model (e.g., Figure~\ref{fig:marmini}b), and a more detailed introduction to GRFs can be found in \citep{izzatullah2023,corm2025, taufik2026}}. The GRF perturbations introduce spatially correlated variations into the velocity models, enabling the particles to cover a broader range of plausible initial solutions while preserving the large-scale background structure. \new{We generate 50 particles, and Figure~\ref{fig:marmpar} shows 9 randomly selected initial particles from the resulting ensemble}, where we can see that although the particles preserve the overall large-scale velocity trend, they exhibit noticeable local variations in both velocity magnitude and spatial structure. These perturbations provide sufficient diversity for posterior exploration while maintaining physically reasonable initial models.

\par
\begin{figure}[htb!]
\centering
\includegraphics[width=1.0\columnwidth]{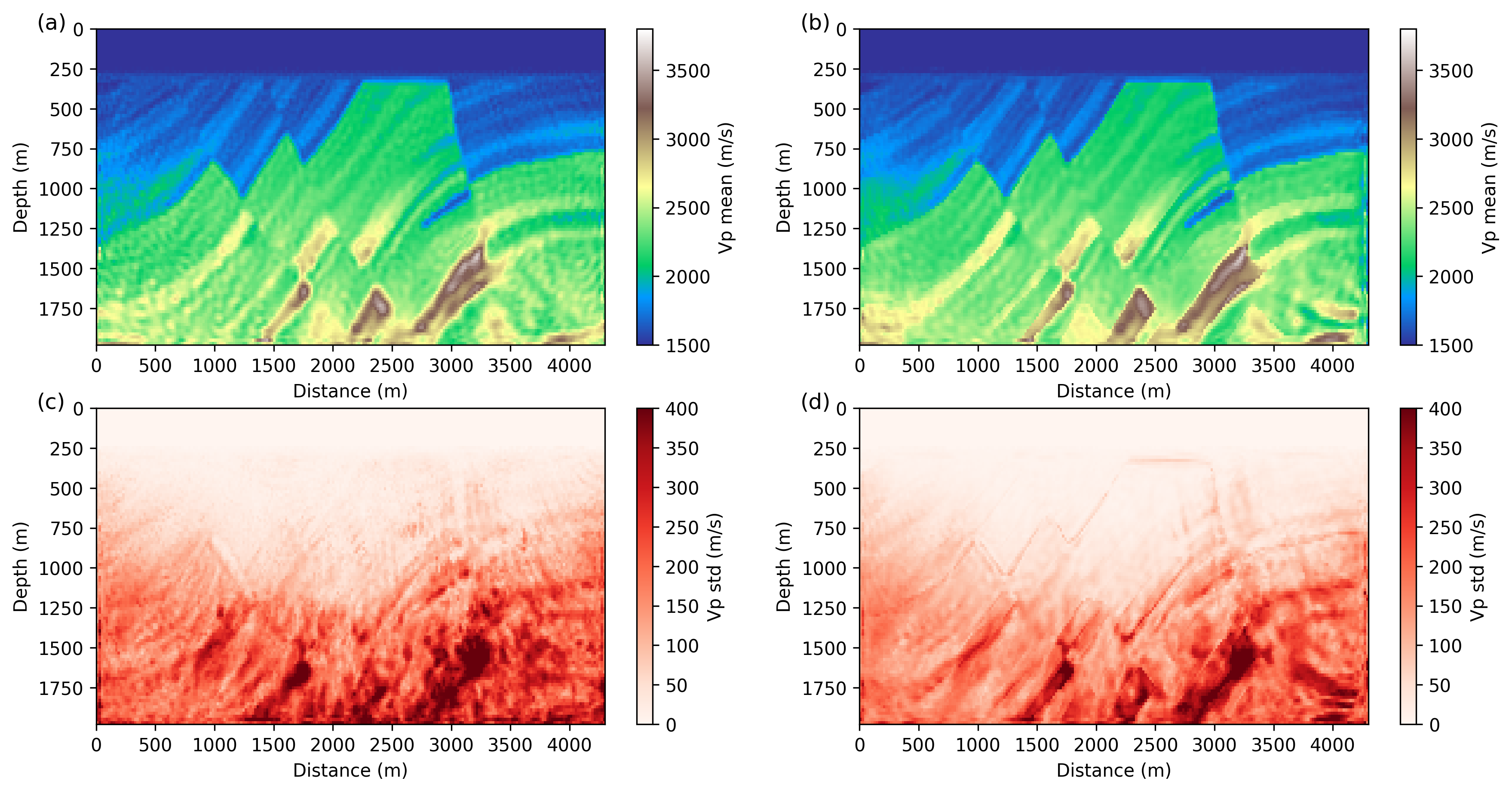}
\caption{Estimated mean value of Marmousi model using (a) classic ASVGD method and (b) \new{diffusion-regularized} ASVGD method. Estimated std value of the Marmousi model using (c) classic ASVGD method and (d) diffusion-regularized ASVGD method. }
\label{fig:marminv}
\end{figure}

 Then we conduct 100 iterations for both the classic ASVGD FWI and the diffusion-regularized ASVGD FWI, with a learning rate of 100. Figure~\ref{fig:marminv} presents the final posterior mean and standard deviation (std) obtained using ASVGD-FWI and diffusion-regularized ASVGD-FWI. Both methods successfully recover the major subsurface structures and provide a reasonable characterization of the associated inversion uncertainty. However, the \new{diffusion-regularized} approach yields improved structural continuity, sharper interfaces, and fewer inversion artifacts, while producing a more spatially concentrated and interpretable uncertainty distribution. \new{To provide a quantitative comparison, Table~\ref{tablemarm} summarizes the Mean Absolute Error (MAE), Structural Similarity Index Measure (SSIM), and model SNR (MSNR), where we can see that compared with classic ASVGD, the proposed method can reduce the error and improve the structural fidelity of the final inversion result.}
\begin{table}
 \caption{\new{Performance statistics of the inverted Marmousi model using classic ASVGD and diffusion-guided ASVGD using different metrics. The symbol $\downarrow$ means the smaller value is better, and $\uparrow$ means the larger value is better. The better results are marked in red.}}
\centering
 \begin{tabular}{@{}ccc}
\toprule
Metric& classic ASVGD& diffusion-guided ASVGD\\\cmidrule(r){1-3}
MAE $\downarrow$ & 94.80 &\textcolor{red}{\textbf{92.68}}\\ 
SSIM $\uparrow$	& 0.69 &\textcolor{red}{\textbf{0.72}}\\ 
MSNR (dB)  $\uparrow$ & 21.19 &\textcolor{red}{\textbf{21.48}} \\ \bottomrule
 \end{tabular}
 \label{tablemarm}
\end{table}
\begin{figure}[htb!]
\centering
\includegraphics[width=1.0\columnwidth]{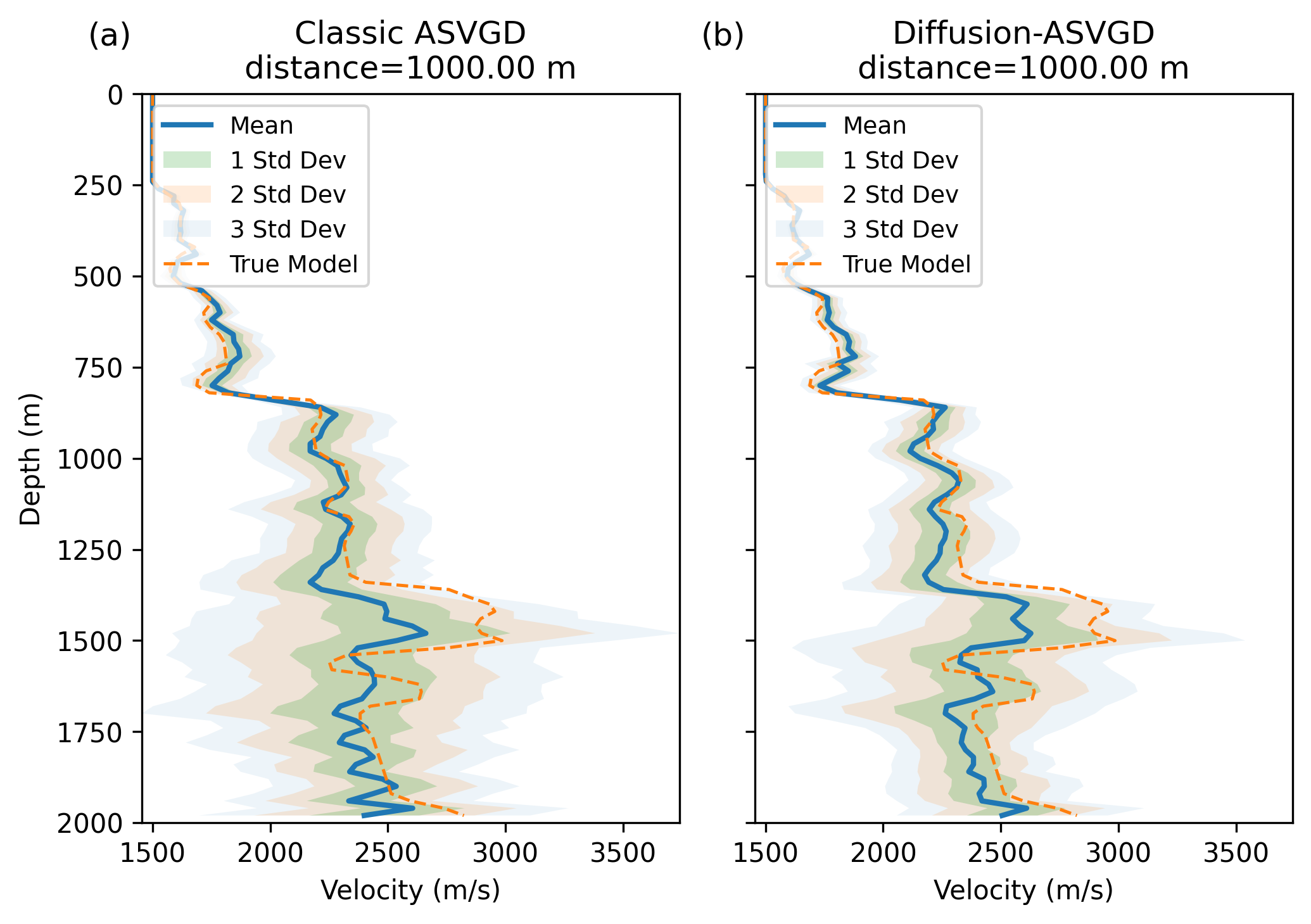}
\caption{\new{Verticle profile comparison of the Marmousi model. (a) Vertical profile showing mean and three confidence intervals for the experiment using classic ASVGD. (b) Vertical profile showing mean and three confidence intervals for the experiment using diffusion-guided ASVGD. }}
\label{fig:marmsing}
\end{figure}
Additionally, the std obtained by conventional ASVGD is widely distributed across the middle and deep parts of the model, suggesting relatively large, spatially dispersed particle variability. In the diffusion-regularized result, the uncertainty is more localized around deep structures, high-contrast interfaces, and structurally complex regions, where the seismic data provide weaker constraints. Meanwhile, the std in relatively smooth and well-resolved regions is reduced. \new{Specifically, Figure~\ref{fig:marmsing} compares vertical velocity profiles obtained using classic ASVGD and diffusion-regularized ASVGD. The diffusion-regularized approach exhibits consistently narrower uncertainty envelopes, particularly in the middle and deeper portions of the model. This suggests that the diffusion prior constrains the particle ensemble more effectively, resulting in a more concentrated posterior distribution and reduced uncertainty while retaining the dominant velocity trends.}

\par
\begin{figure}[htb!]
\centering
\includegraphics[width=1.0\columnwidth]{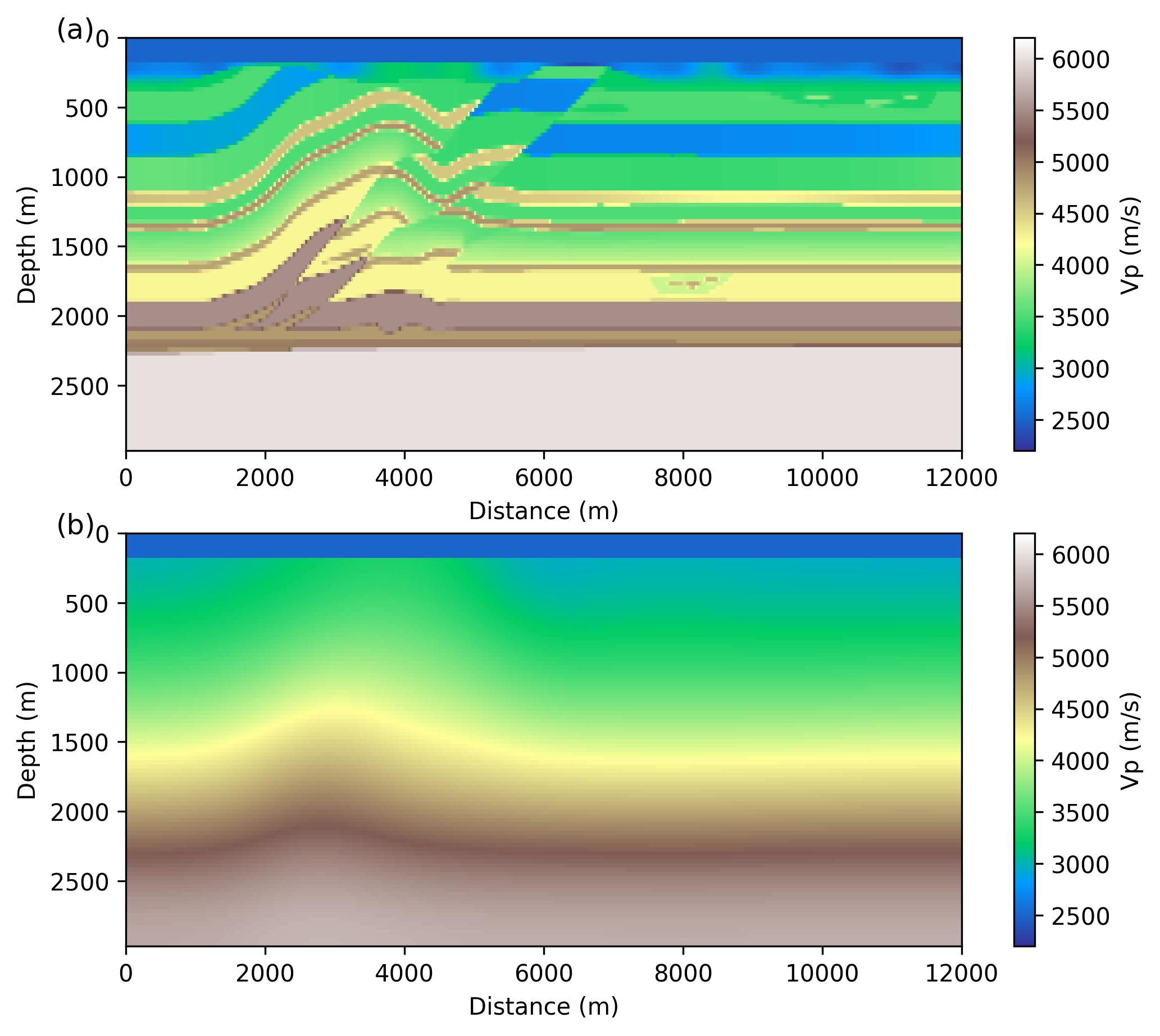}
\caption{(a) Overthrust model (ground truth). (b) Initial Overthrust model used for inversion. }
\label{fig:overini}
\end{figure}

\begin{figure}[htb!]
\centering
\includegraphics[width=1.0\columnwidth]{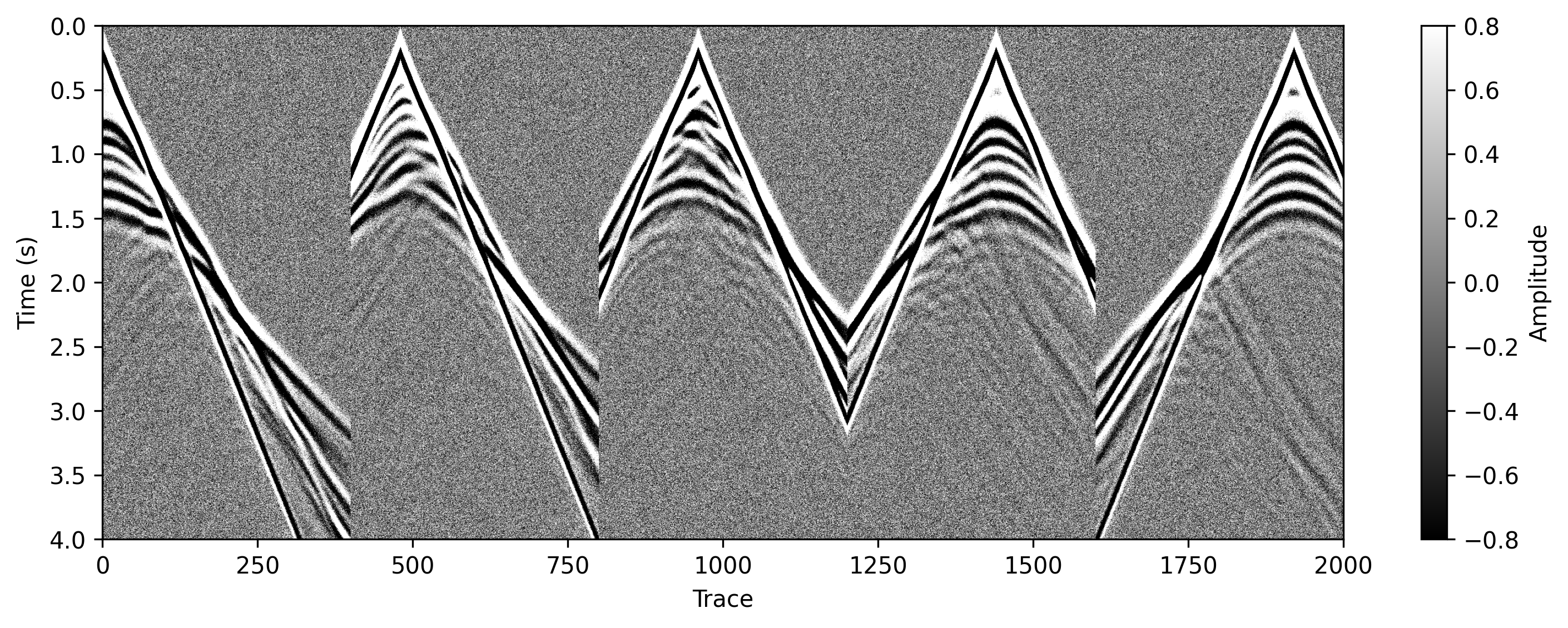}
\caption{Observed data generated using real Overthrust model (DSNR=16.9 dB).  }
\label{fig:overdata}
\end{figure}

\begin{figure}[htb!]
\centering
\includegraphics[width=1.0\columnwidth]{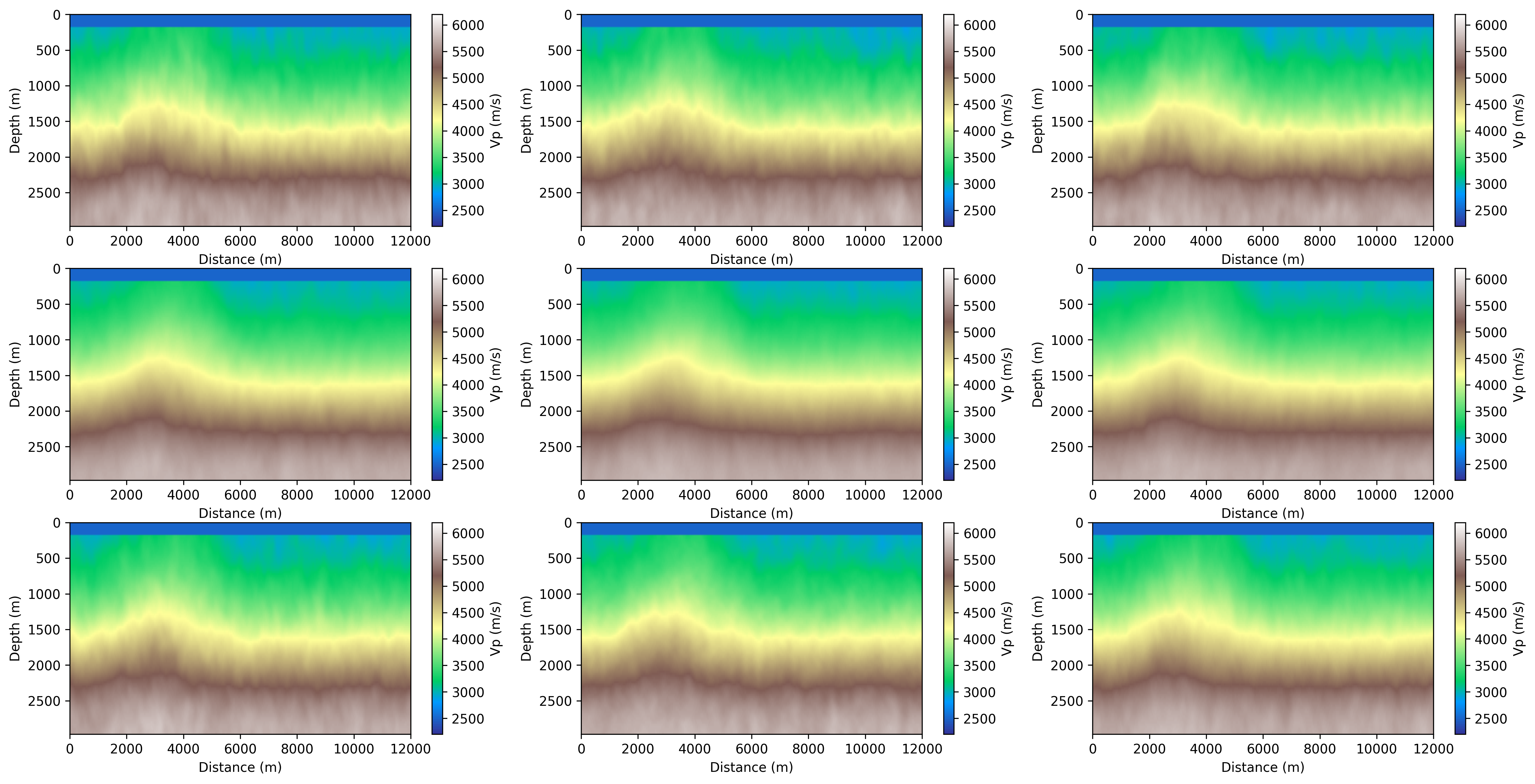}
\caption{Nine randomly selected initial particles generated by perturbing the smoothed Overthrust model with GRFs.}
\label{fig:overpar}
\end{figure}

\begin{figure}[htb!]
\centering
\includegraphics[width=1.0\columnwidth]{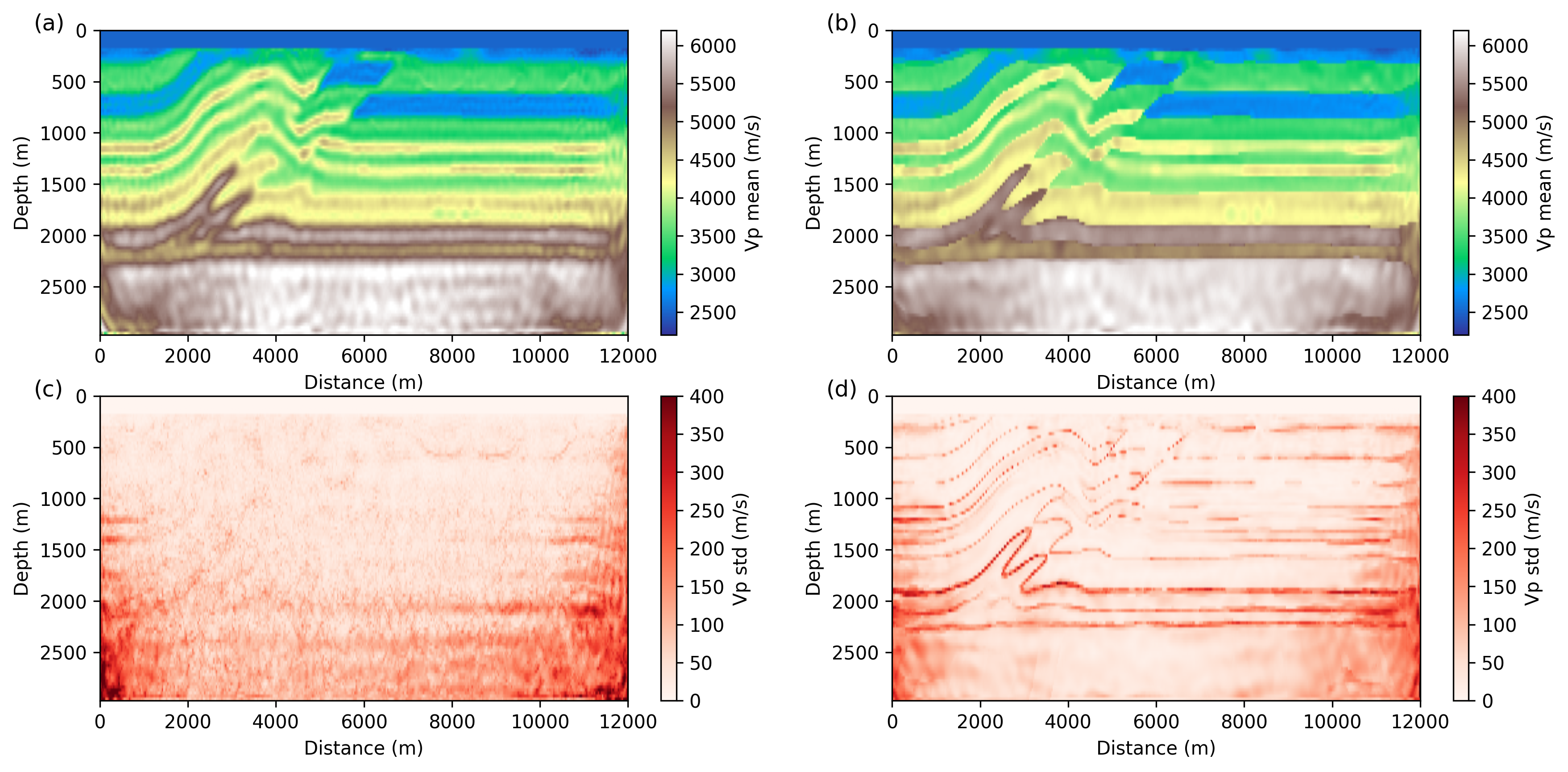}
\caption{Estimated mean value of the Overthrust model using (a) the classic ASVGD method and (b) the diffusion-guided ASVGD method. Estimated standard deviation value of the Overthrust model using (c) the classic ASVGD method and (d) the diffusion-guided ASVGD method. }
\label{fig:overinv}
\end{figure}

\begin{figure}[htb!]
\centering
\includegraphics[width=1.0\columnwidth]{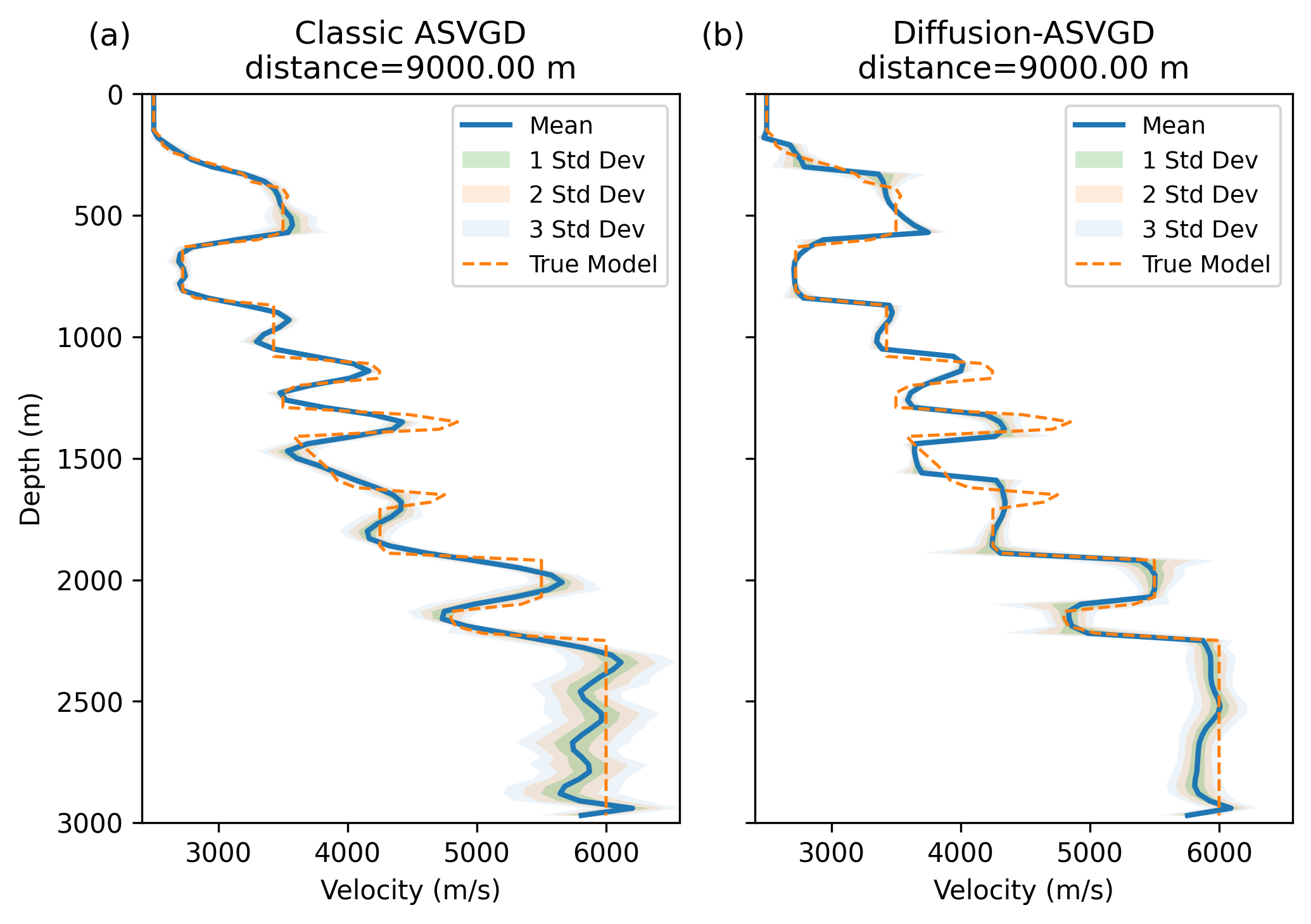}
\caption{\new{Verticle profile comparison of the Overthrust model. (a) Vertical profile showing mean and three confidence intervals for the experiment using classic ASVGD. (b) Vertical profile showing mean and three confidence intervals for the experiment using diffusion-guided ASVGD. }}
\label{fig:oversing}
\end{figure}

 Next, we use the Overthrust model to further evaluate the superiority of the diffusion-regularized ASVGD method. Figure~\ref{fig:overini} presents the ground-truth and initial Overthrust velocity models used in the inversion. \new{The computational domain is discretized into a 100 $\times$ 401 grid with a spatial interval of 30 m. A 5 Hz Ricker wavelet sampled at a temporal interval of 2 ms is used as the source signature. We set four absorbing boundary conditions, and 10 shots are uniformly distributed along the surface.} Figure~\ref{fig:overdata} shows 5 observed shots corresponding to a real Overthrust model, where Gaussian noise is added, and the DSNR is about 16.9 dB. Similar to the previous example, we generate 50 initial particles by perturbing the initial model shown in Figure~\ref{fig:overini}b with GRFs. Figure~\ref{fig:overpar} presents nine representative particles from the resulting ensemble. These particles remain relatively smooth and preserve similar large-scale, low-frequency velocity characteristics, while exhibiting sufficient spatial variability to support posterior exploration. Based on the 50 initial particles generated, we also perform 100 inversion iterations with a learning rate of 100 for both classic ASVGD and diffusion-regularized ASVGD. Figure~\ref{fig:overinv} compares the posterior mean and std obtained using conventional ASVGD and diffusion-regularized ASVGD. Although both methods recover the major structural features of the Overthrust model, the diffusion-regularized result in Figure~\ref{fig:overinv}b exhibits sharper layer boundaries, improved structural continuity, and fewer spatially scattered artifacts than the conventional ASVGD result shown in Figure~\ref{fig:overinv}a. In particular, the folded structures in the shallow and middle parts of the model, as well as the major high-velocity interfaces at greater depths, are more clearly delineated after introducing the diffusion prior. This indicates that the learned structural prior effectively guides the particles toward geologically plausible regions of the model space.
Moreover, the std produced by conventional ASVGD in Figure~\ref{fig:overinv}c is broadly distributed, especially in the deeper part of the model, suggesting relatively diffuse particle variability. By contrast, the uncertainty shown in Figure~\ref{fig:overinv}d is more localized around major interfaces, folded structures, and high-contrast boundaries, where limited illumination and strong structural complexity make the model parameters more difficult to constrain. Such a localized uncertainty pattern is more closely associated with structurally complex regions and is therefore more interpretable from a geological perspective. \new{Moreover, Table~\ref{tableover} shows the quantitative comparison corresponding to  the Overthrust model, where we can see that with the help of diffusion prior, the proposed method can lead to a lower residual and reach a better result with higher structure fidelity.} \new{In detail, Figure~\ref{fig:oversing} shows the vertical profile comparison corresponding to Overthrust model. We can see that diffusion-ASVGD provides a posterior mean that more closely follows the true velocity profile. Some visible improvements become more pronounced below approximately 2 km depth, where classic ASVGD shows broader uncertainty intervals and stronger fluctuations, while diffusion-ASVGD maintains a more concentrated and stable posterior distribution.} Overall, the diffusion-regularized ASVGD method provides better structural fidelity, suppresses nonphysical fluctuations, and yields a more spatially meaningful characterization of posterior uncertainty.
\begin{table}
 \caption{\new{Performance statistics of the inverted Overthrust model using classic ASVGD and diffusion-guided ASVGD using different metrics. The symbol $\downarrow$ means the smaller value is better, and $\uparrow$ means the larger value is better. The better results are marked in red.}}
\centering
 \begin{tabular}{@{}ccc}
\toprule
Metric& classic ASVGD& diffusion-guided ASVGD\\\cmidrule(r){1-3}
MAE $\downarrow$ & 173 &\textcolor{red}{\textbf{153}}\\ 
SSIM $\uparrow$	& 0.62 &\textcolor{red}{\textbf{0.64}}\\ 
SNR (dB)  $\uparrow$ & 24.7 &\textcolor{red}{\textbf{25.3}} \\ \bottomrule
 \end{tabular}
 \label{tableover}
\end{table}

\subsection{Field data}
\par
\begin{figure}[htb!]
\centering
\includegraphics[width=1.0\columnwidth]{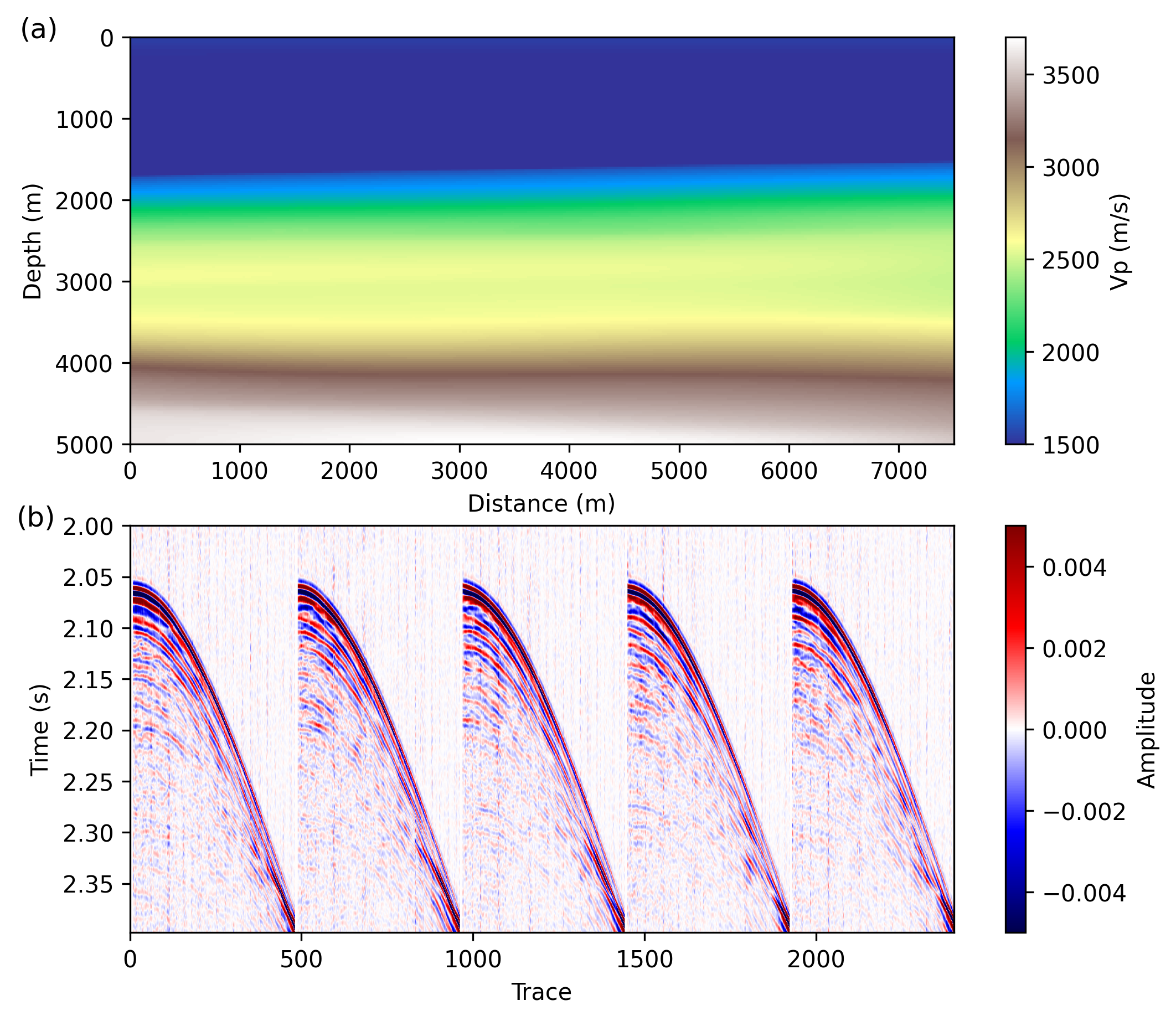}
\caption{(a) Initial model corresponding to the field data. (b) Partial field shot gathers. }
\label{fig:fldini}
\end{figure}

\begin{figure}[htb!]
\centering
\includegraphics[width=1.0\columnwidth]{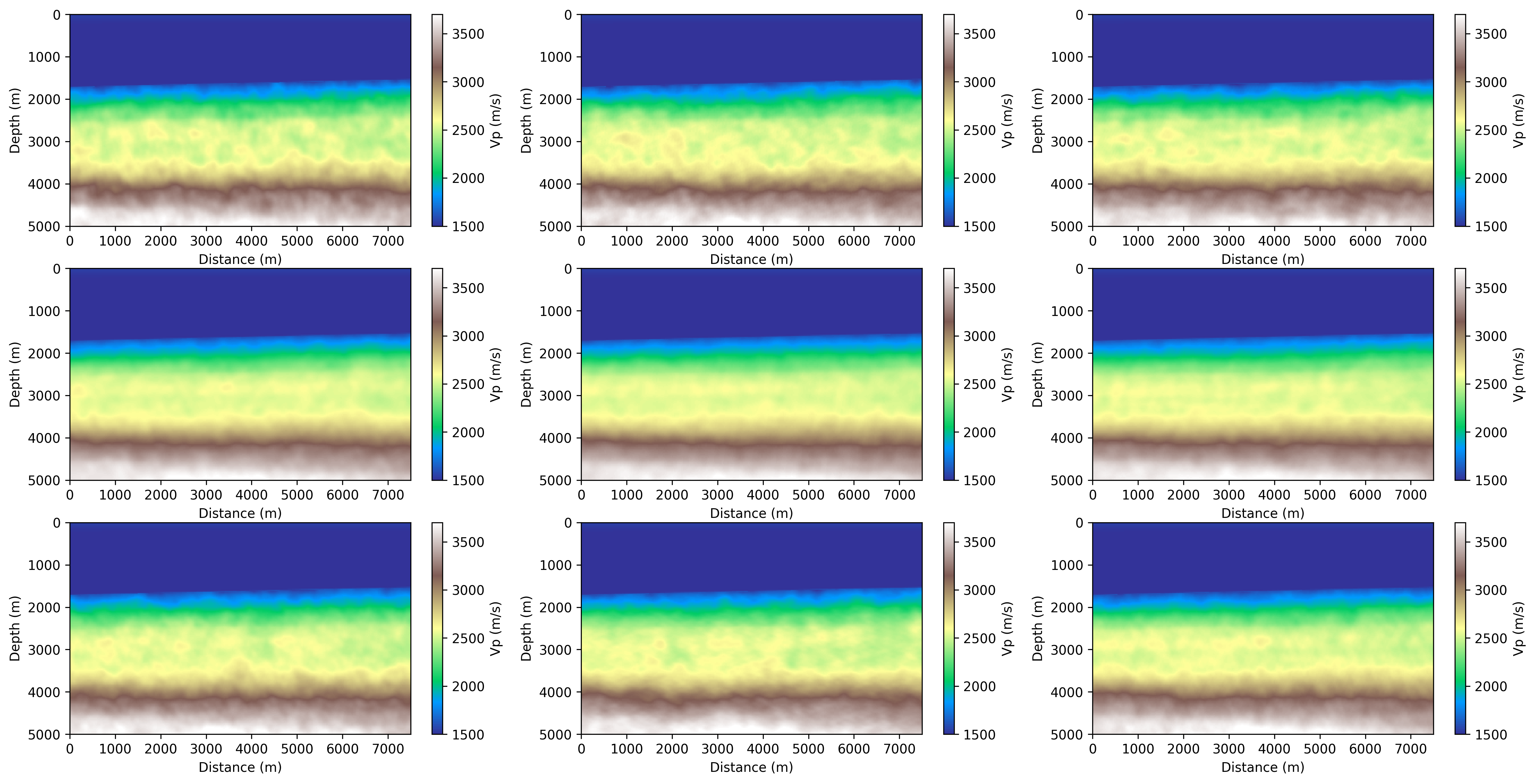}
\caption{Nine randomly selected initial particles for the field data case, generated by perturbing the initial model using GRFs. }
\label{fig:fldpar}
\end{figure}

\begin{figure}[htb!]
\centering
\includegraphics[width=1.0\columnwidth]{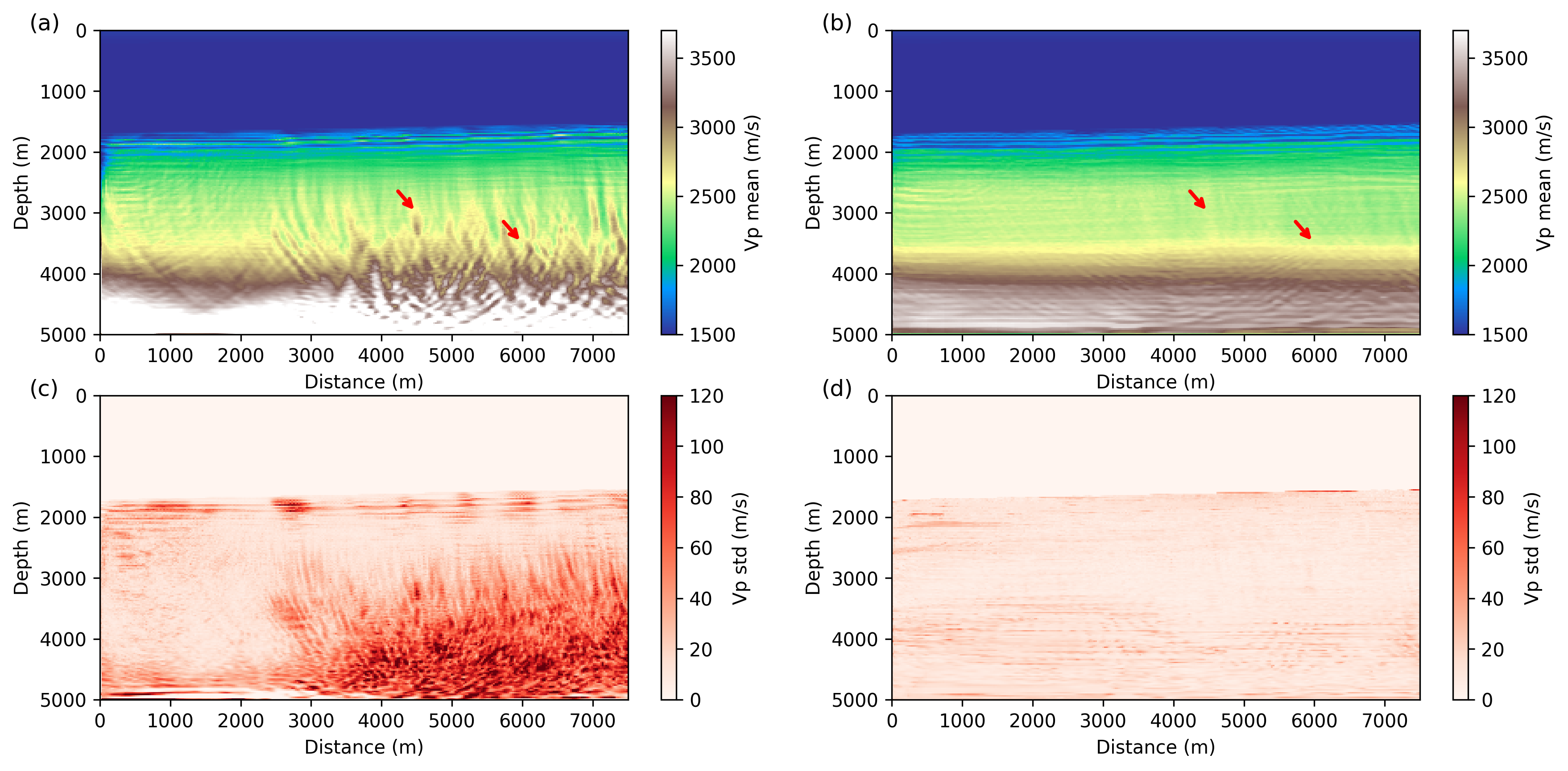}
\caption{Field-data inversion results. Estimated mean velocity models obtained using (a) the classic ASVGD method and (b) the diffusion-regularized ASVGD method. Estimated standard deviation maps obtained using (c) the classic ASVGD method and (d) the diffusion-regularized ASVGD method. The red arrows highlight the artifacts may be caused by insufficient illumination, acquisition limitations, or noise in the field data. }
\label{fig:fldinv}
\end{figure}

 We then apply the proposed method to a field dataset acquired in offshore China to assess its feasibility in practice. Figure~\ref{fig:fldini}a presents the initial velocity model corresponding to the field data. The model is discretized into 201 $\times$ 301 grid points with a spatial sampling interval of 25 m. \new{The acquisition geometry consists of 40 shots, with the seismic data band-pass filtered between 5 and 15 Hz. The sources are spaced at 200 m intervals, and the temporal sampling interval is 2 ms}. As shown in Figure~\ref{fig:fldini}b, each shot gather contains 240 traces with a receiver spacing of 25 m. Like the previous synthetic example, we generate 50 particles by adding GRFs to the initial model shown in Figure~\ref{fig:fldini}, and the partially generated particles are displayed in Figure~\ref{fig:fldpar}. From Figure~\ref{fig:fldpar}, we can see that the background velocity structures remain largely consistent across particles, whereas the local perturbations show clear diversity. Based on the generated particles, we then perform 30 iterations for both the classic ASVGD and the diffusion-regularized ASVGD FWI, using a constant learning rate of 30. Figure~\ref{fig:fldinv} depicts the final estimated mean value and uncertainty distribution of the field data, showing that, compared with the classic ASVGD result, the diffusion-regularized ASVGD method produces a more geologically consistent velocity model with smoother, more continuous background structures. As indicated by the red arrows, the classic ASVGD result shows several localized anomalous perturbations, likely due to insufficient illumination, acquisition limitations, or noise in the field data. These artifacts result in less geologically plausible velocity structures and greater uncertainty in the corresponding std map. In contrast, the diffusion-regularized ASVGD method suppresses these non-geological anomalies and produces a more coherent, structurally plausible velocity model. These observations suggest that the diffusion prior effectively suppresses non-geological artifacts and guides the particles toward a more reliable posterior distribution.

\begin{figure}[htb!]
\centering
\includegraphics[width=1.0\columnwidth]{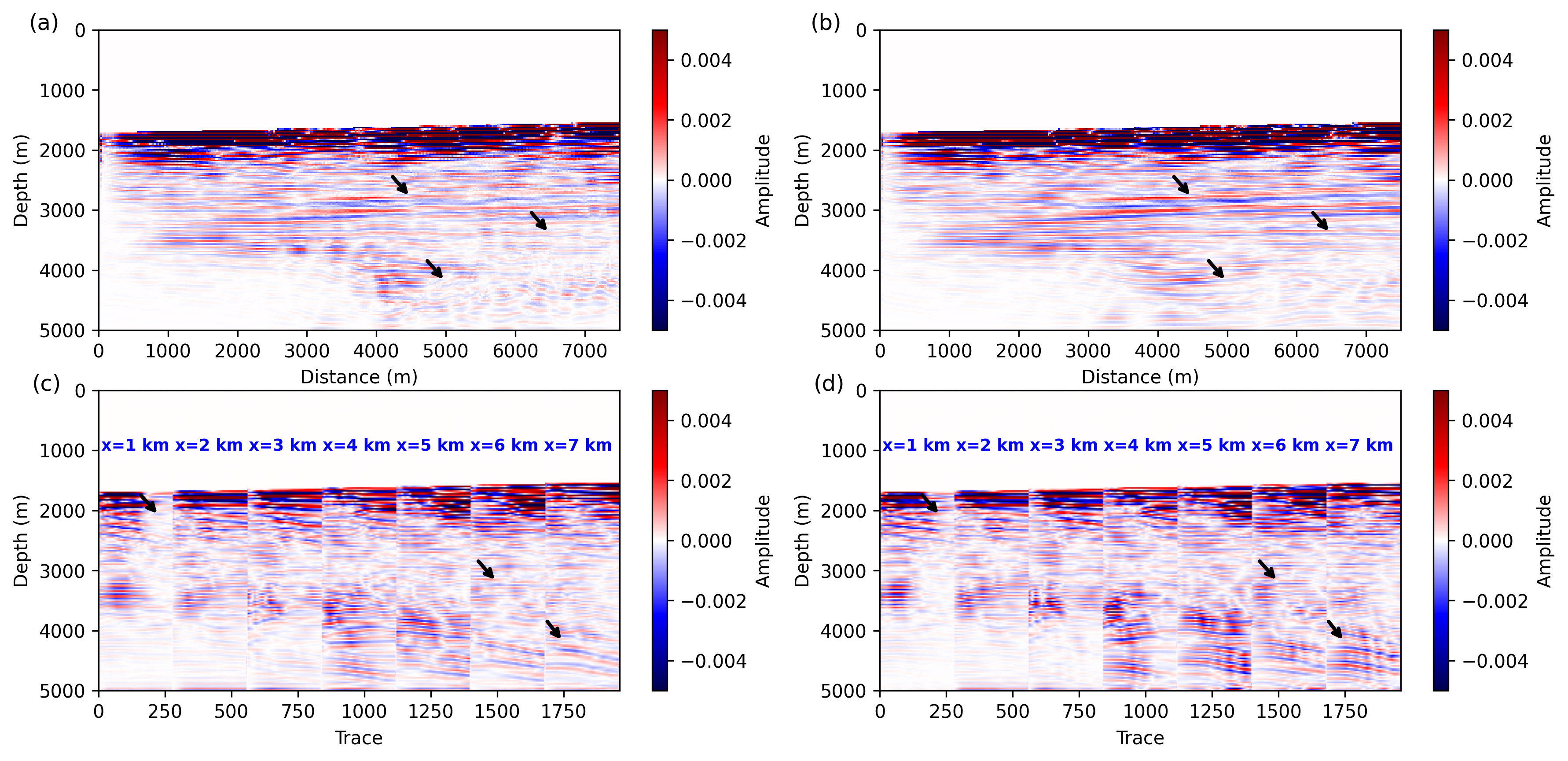}
\caption{Migration profile obtained using the velocity model of (a) classic ASVGD method and (b) \old{diffusion-guided}\new{diffusion-reggularized} ASVGD method. (c) ADCIG at different locations corresponding to Fig~\ref{fig:fldrtm}a. (d) ADCIG at different locations corresponding to Fig~\ref{fig:fldrtm}b. The black arrows highlight the visible improvements in the image profile and ADCIG gathers, indicating enhanced image resolution. }
\label{fig:fldrtm}
\end{figure}

To evaluate the reliability of the inversion results, we then perform reverse-time migration (RTM) and extract the angle-domain common image gather (ADCIG) for further intuitive comparison. Figure~\ref{fig:fldrtm} shows the final migration profiles and ADCIG at different locations. Obviously, compared with the classic ASVGD result, the diffusion-regularized ASVGD result produces a better-focused RTM image with more continuous reflectors (e.g., marked by the black arrows). The corresponding ADCIGs also show flatter events and weaker residual moveout, indicating improved kinematic consistency of the inverted mean velocity model. 
Moreover, we can see that the event energy is more focused in Figure~\ref{fig:fldrtm}d. These observations suggest that the diffusion prior helps suppress non-geological velocity anomalies caused by insufficient illumination or field-data noise, thereby improving the structural plausibility and imaging reliability of conventional ASVGD FWI.

\section{Discussion}
In this section, we will interpret the potential mechanism of diffusion prior and conduct additional experiments to evaluate the robustness and the limitations of the proposed method. All the experiments are based on the Overthrust model shown in Figure~\ref{fig:overini}.
\subsection{Role of the diffusion prior in structural information incorporation}
\par
\begin{figure}[htb!]
\centering
\includegraphics[width=1.0\columnwidth]{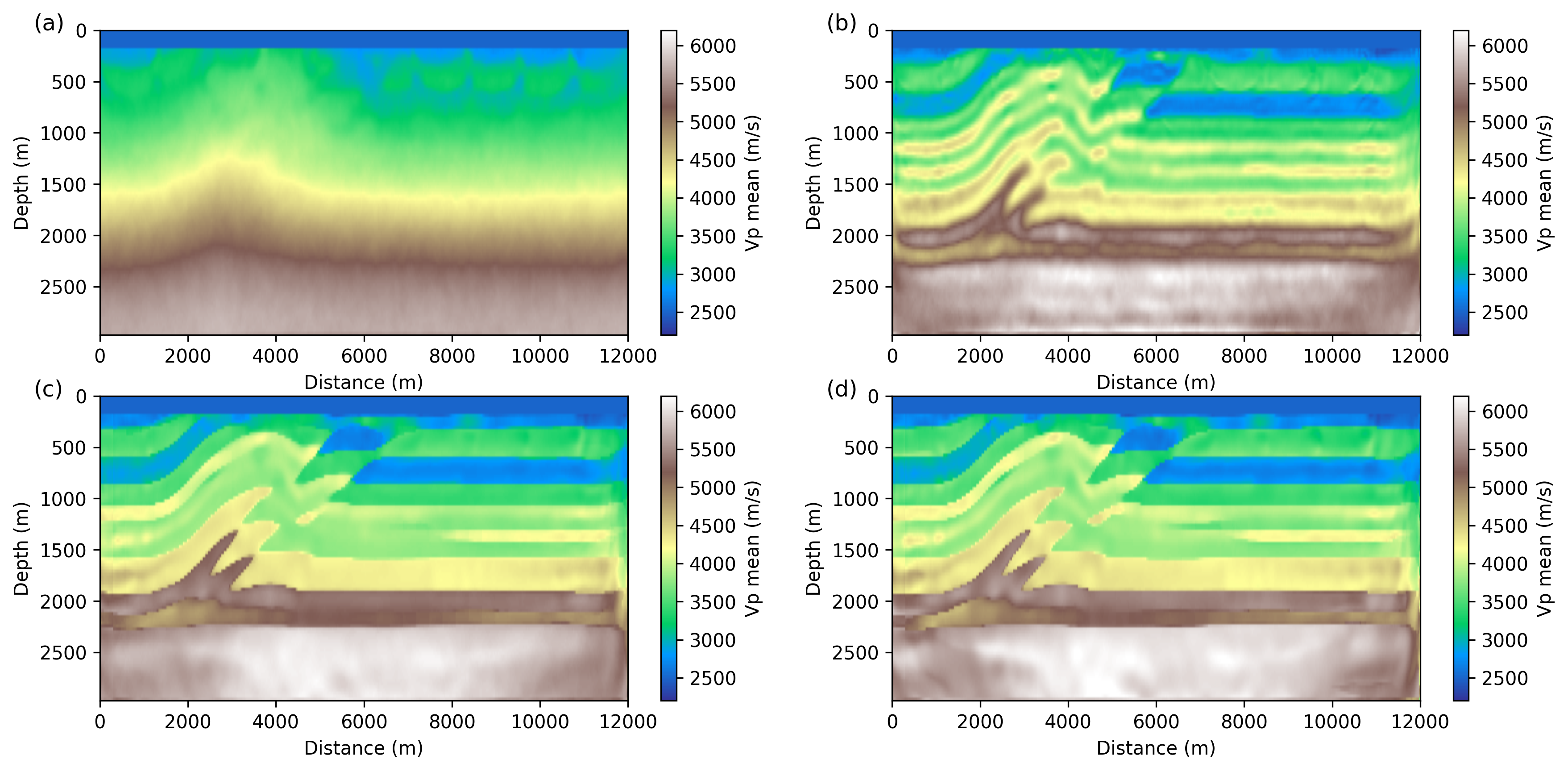}
\caption{Evolution of the velocity mean at different iterations: (a) 5, (b) 25, (c) 45, and (d) 85 iterations. The results demonstrate that the inversion process progressively incorporates structural information, evolving from a smooth large-scale background model to a geologically consistent model with coherent layers, sharper velocity contrasts, and more detailed fault-related structures. }
\label{fig:overevo}
\end{figure}

\begin{figure}[htb!]
\centering
\includegraphics[width=1.0\columnwidth]{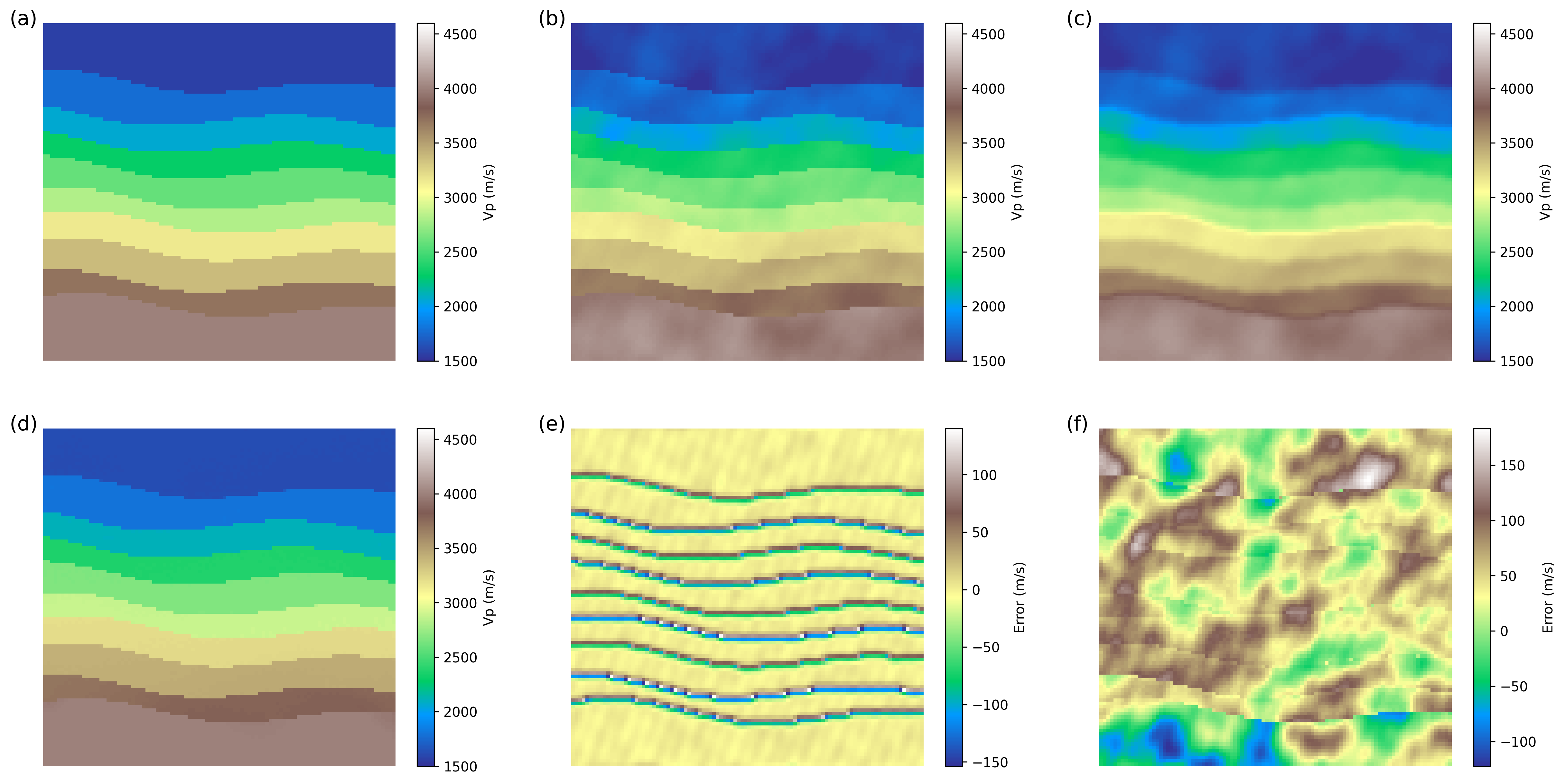}
\caption{\new{Comparison of Gaussian smoothing and diffusion-based regularization for suppressing GRF-induced perturbations. (a) Reference velocity model. (b) GRF-perturbed velocity model. (c) Result after conventional Gaussian smoothing. (d) Result after diffusion-based regularization. (e) Difference between the Gaussian-smoothed model and the reference model. (f) Difference between the diffusion-regularized model and the reference model. The diffusion prior more effectively suppresses GRF-induced perturbations while preserving the layer boundaries and structural features. }}
\label{fig:dmgrf}
\end{figure}

 We first show an intuitive example to help understand the potential mechanism of diffusion prior and its role of structural information incorporation. Figure~\ref{fig:overevo} illustrates the progressive incorporation of structural information in diffusion-regularized ASVGD-FWI. In the early iterations, the recovered model primarily captures the large-scale velocity trend, whereas the layered interfaces and complex structural features remain relatively blurred. This behavior indicates that the inversion is initially dominated by the data-driven update, allowing the model to account for the observed data before more detailed structural constraints are introduced. With increasing iterations, fault-related structures, curved stratigraphic layers, and sharp velocity contrasts become increasingly pronounced and better organized. The diffusion prior therefore acts as a progressive structural regularizer, enhancing geologically meaningful features while suppressing unstable artifacts. At 85 iterations, the model preserves the large-scale background trend while recovering finer structural details, demonstrating that the diffusion prior and the data-driven SVGD update work together to produce a geologically consistent and structurally informative inversion result. \new{Furthermore, since the initial particles in SVGD are generated by adding GRF perturbations to the initial velocity model, we further evaluate the effectiveness of the learned diffusion prior in suppressing such GRF-induced perturbations. The results are shown in Figure~\ref{fig:dmgrf}. In this experiment, we randomly select a velocity model with folded layers from the GeoFWI dataset, as shown in Figure~\ref{fig:dmgrf}(a), and add a GRF perturbation to obtain the perturbed model shown in Figure~\ref{fig:dmgrf}(b). Figures~\ref{fig:dmgrf}(c)-(f) present the regularized results and the corresponding error maps obtained using conventional Gaussian smoothing (e.g., variance is set to 1) and diffusion-based regularization, respectively. Compared with conventional Gaussian smoothing, the learned diffusion prior more effectively suppresses the GRF-induced perturbations while preserving sharp layer boundaries and structural details. In contrast, Gaussian smoothing tends to blur the interfaces and introduces larger deviations from the reference model, thereby compromising structural fidelity.}

\subsection{Influence of the number of initial particles}
\par
\begin{figure}[htb!]
\centering
\includegraphics[width=1.0\columnwidth]{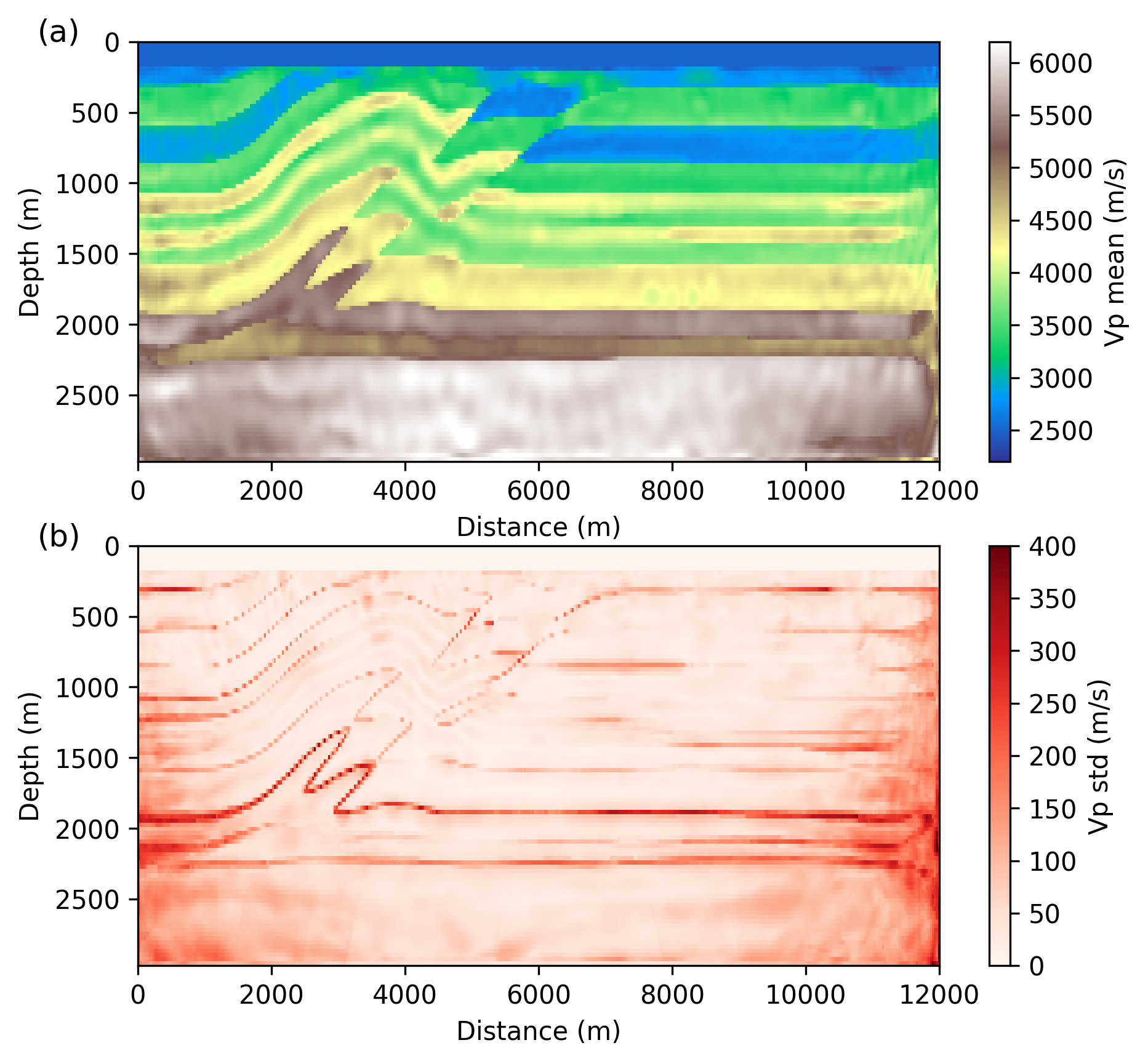}
\caption{(a) Estimated velocity mean value using 100 different initial particles. (b) Estimated std map using 100 different initial particles.  }
\label{fig:overnum}
\end{figure}

 We then discuss the number of initial particles on the final results. Under the same configuration, we increase the number of particles to 100 and keep the other parameters unchanged for inversion. Figure~\ref{fig:overnum} shows the final estimated mean value and std distribution after 100 iterations, where we can see that the results obtained with 100 particles show a similar posterior mean model to that obtained with 50 particles. This indicates that 50 particles are already sufficient to capture the dominant features of the posterior distribution under the guidance of the diffusion prior. Increasing the number of particles to 100 does not lead to a substantial improvement in the mean velocity model. However, the uncertainty estimation becomes more stable when more particles are used. Compared with the result obtained with 50 particles, the std map from 100 particles shows more continuous and statistically stable uncertainty patterns, especially around complex structures and deep interfaces. This is because a larger number of particles provides a better approximation of the posterior distribution and reduces the influence of random fluctuations from individual particles.
Therefore, increasing the number of particles mainly improves the stability of posterior uncertainty estimation, whereas the improvement in the posterior mean model is relatively limited. Meanwhile, the computational cost increases almost proportionally with the number of particles. \new{In our experiments, we use an NVIDIA GeForce RTX 5090 GPU. The inversion with 50 particles requires approximately 20 minutes, whereas increasing the particle number to 100 roughly doubles the computational time to about 40 minutes}. Thus, 50 particles provide a reasonable balance between computational efficiency and inversion stability.

\subsection{Influence of noise level on the final uncertainty evaluation}
\begin{figure}[htb!]
\centering
\includegraphics[width=1.0\columnwidth]{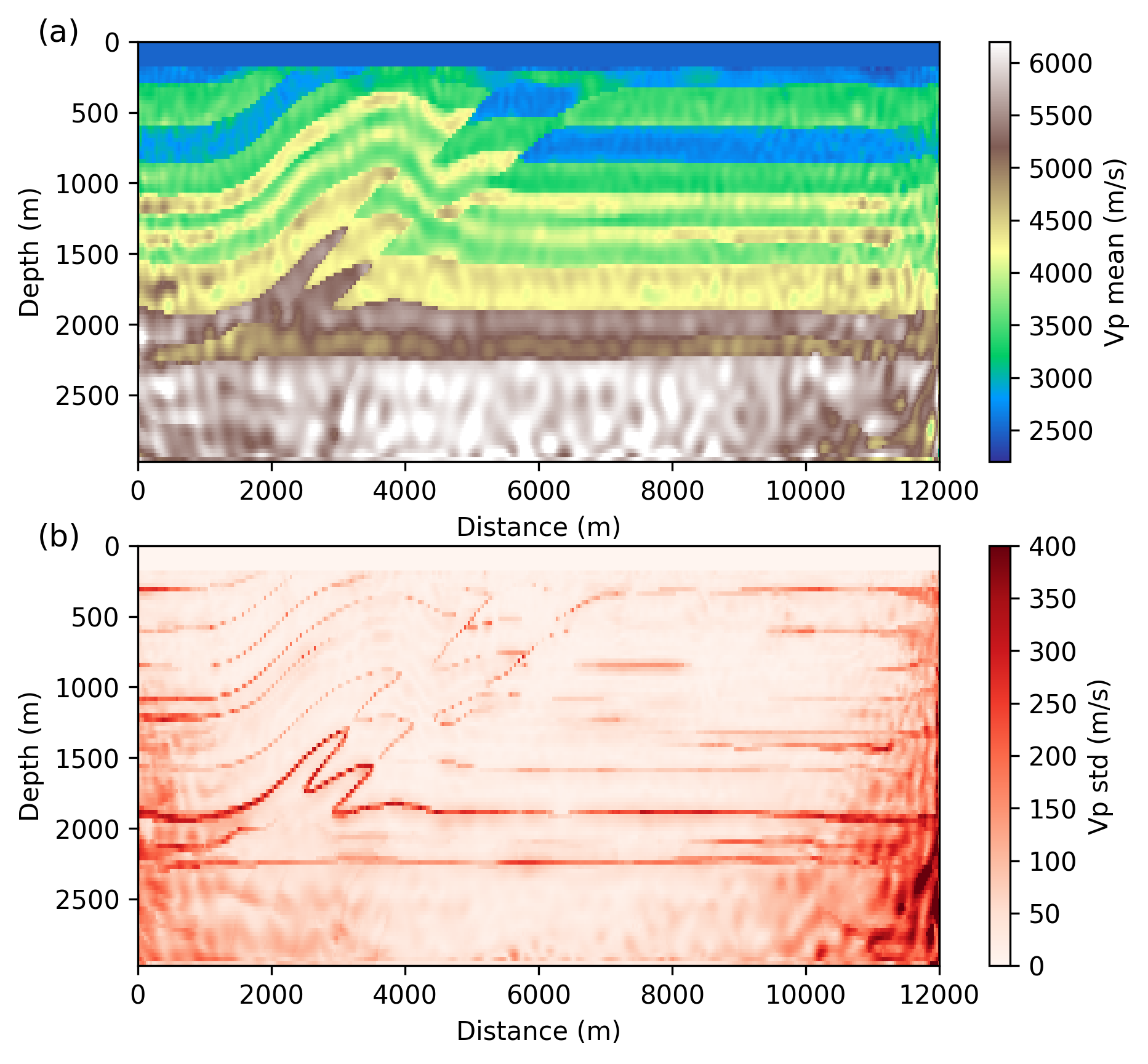}
\caption{(a) Estimated velocity mean value under a DSNR of 2 dB. (b) Estimated std map under a DSNR of 2 dB.  }
\label{fig:oversnr}
\end{figure}

Next, we investigate the influence of noise on the final inversion results by adding stronger noise to the generated shot gathers, yielding a DSNR of 2 dB. In this experiment, the same 50 initial particles are used, and all other inversion parameters are kept unchanged. The final inversion results are shown in Figure~\ref{fig:oversnr}. As expected, the proposed diffusion-regularized ASVGD method can still recover the main velocity structures, indicating its robustness under strong noise contamination. However, compared with the result shown in Figure~\ref{fig:overinv}b, the posterior mean model becomes less continuous and contains more local perturbations, especially in the deep region and near the model boundaries. This is because strong noise contaminates the data-misfit gradient, making particle evolution less stable. Even under this low-DSNR condition, the diffusion-regularized inversion still preserves the main geological structures and converges to a relatively plausible solution. This indicates that the diffusion prior helps stabilize the ASVGD particle evolution by providing additional geological constraints beyond the noisy data likelihood. Although inversion uncertainty inevitably increases with higher noise levels, the proposed method shows improved robustness to noise and can partially mitigate the adverse effects of noisy observations.

\subsection{Key factors in diffusion regularization and computational cost analysis }
\new{Finally, we compare the computational costs of the classic ASVGD method and the diffusion-regularized ASVGD method. Taking the Overthrust example shown in Figure~\ref{fig:overinv} as an illustration, the total computational cost for classic ASVGD and diffusion-regularized ASVGD are approximately 870 s and 964 s, respectively. This comparison indicates that incorporating the diffusion prior introduces only a modest additional computational overhead while providing the benefits of diffusion-based regularization. Additionally, the choice of the starting timestep and shift size for the diffusion regularization also play an important role. In general, an earlier starting timestep (e.g., $t<100$)) results in a more deterministic denoising process, whereas a later starting timestep (e.g., t is closer to 1000) introduces greater stochasticity and uncertainty. As the inversion progresses and the estimated model becomes increasingly accurate, an earlier starting timestep is therefore preferred to reduce unnecessary perturbations and improve inversion accuracy. In our experiment, we set shift size to 10 to reduce the boundary influence related with patching scheme and more discussion can be found in \citep{lic2026}. More importantly, the training dataset used for the diffusion model can also significantly influence the final performance. Constructing geologically plausible and sufficiently diverse training datasets is essential for improving generalizability, particularly in field-data applications, because a broader range of samples enables the network to implicitly learn richer structural and geological features. Once trained, the diffusion prior can then be incorporated into the inversion as an adaptive, data-driven constraint without any extra parameter selection and local slope estimation.  }

\section {Conclusion}
In this paper, we develop a novel diffusion-regularized ASVGD method to improve the stability and structural reliability of particle-based Bayesian full-waveform inversion. The proposed method incorporates a learned diffusion prior into the ASVGD particle evolution, allowing the inversion to be guided not only by the data likelihood but also by geological prior information learned from a diverse and representative dataset of velocity models. In this way, the diffusion guidance provides an effective regularization mechanism for suppressing non-geological perturbations while preserving large-scale background velocity trends and structurally coherent features. Numerical and field data examples validate that the integration of diffusion-based geological priors with ASVGD-FWI, which enhances conventional particle-based Bayesian inversion by jointly improving posterior estimation, uncertainty characterization, and imaging consistency, which provides a promising approach for mitigating noise- and illumination-related artifacts in challenging FWI applications, especially when reliable prior information is needed to stabilize the inversion of field data.

\section{Acknowledgements}
 The authors would like to thank Texas Consortium of Computational Seismology (TCCS), the Texas Seismological Network and Seismology Research (TexNet), and the State of Texas, which provided financial support for this publication under the University of Texas at Austin award \# 201503664. 

\section{Conflict of Interest}
The authors declare no conflicts of interest relevant to this study.

\section{Open Research Section}
The open-source repository of diffusion-guided ASVGD FWI, including codes, and examples, can be accessed via \citep{lidmsvgd}.

\bibliographystyle{unsrt}
\bibliography{my}
\end{document}